\documentclass[letterpaper]{article} 
\usepackage{aaai2026}  
\usepackage{times}  
\usepackage{helvet}  
\usepackage{courier}  
\usepackage[hyphens]{url}  
\usepackage{graphicx} 
\usepackage{natbib}  
\usepackage{caption} 
\usepackage{url}            
\usepackage{booktabs}       
\usepackage{amsfonts}       
\usepackage{nicefrac}       
\usepackage{microtype}      
\usepackage{xcolor}  
\usepackage{bm}       
\usepackage{amsthm}
\usepackage{amsmath}
\usepackage{subcaption}
\usepackage[ruled]{algorithm2e}
\usepackage{algorithmicx}
\usepackage{algpseudocode}
\newtheorem{proposition}{Proposition}
\newtheorem{definition}{Definition}
\definecolor{SkyBlue}{RGB}{0, 0, 0}

\usepackage{newfloat}
\usepackage{listings}
\DeclareCaptionStyle{ruled}{labelfont=normalfont,labelsep=colon,strut=off} 
\title{\includegraphics[width=0.9cm]{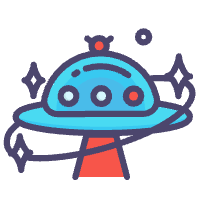} DragNeXt: Rethinking Drag-Based Image Editing}
\author {
    Yuan Zhou\textsuperscript{\rm 1},
    Junbao Zhou\textsuperscript{\rm 1},
    Qingshan Xu\textsuperscript{\rm 1}\thanks{Corresponding author.},
    Kesen Zhao\textsuperscript{\rm 1},
    Yuxuan Wang\textsuperscript{\rm 1},
    Hao Fei\textsuperscript{\rm 2},\\
    Richang Hong\textsuperscript{\rm 3},
    Hanwang Zhang\textsuperscript{\rm 1}
}
\affiliations {
    \textsuperscript{\rm 1}Nanyang Technological University\\
    \textsuperscript{\rm 2}National University of Singapore\\
    \textsuperscript{\rm 3}Hefei University of Technology\\
}

\usepackage{bibentry}

\begin{document}

\maketitle

\begin{abstract}
Drag-Based Image Editing (DBIE), which allows users to manipulate images by directly dragging objects within them, has recently attracted much attention from the community. However, it faces two key challenges: (\emph{\textcolor{black}{i}}) point-based drag is often highly ambiguous and difficult to align with user intentions; (\emph{\textcolor{black}{ii}}) current DBIE methods primarily rely on alternating between motion supervision and point tracking, which is not only cumbersome but also fails to produce high-quality results. These limitations motivate us to explore DBIE from a new perspective---unifying it as  a Latent Region Optimization (LRO) problem that aims to use region-level geometric transformations to optimize latent code to realize drag manipulation. Thus, by specifying the areas and types of geometric transformations, we can effectively address the ambiguity issue. We also propose a simple yet effective editing framework, dubbed \textbf{\textcolor{SkyBlue}{DragNeXt}}. It solves LRO through Progressive Backward Self-Intervention (PBSI), simplifying the overall procedure of the alternating workflow while further enhancing quality by fully leveraging region-level structure information and progressive guidance from intermediate drag states. We validate \textbf{\textcolor{SkyBlue}{DragNeXt}} on our NextBench, and extensive experiments demonstrate that our proposed method can significantly outperform existing approaches. 
\end{abstract}

\begin{links}
\link{Code}{https://github.com/zhouyuan888888/DragNeXt}
\link{Extended version}{https://arxiv.org/pdf/2506.07611}
\end{links}

\section{Introduction}
\label{sec:intro}

Diffusion models \cite{rombach2022high,dhariwal2021diffusion} have made remarkable progress in the field of text-to-image generation, serving as foundational models for a wide range of generative tasks, such as image super-resolution \cite{wu2024seesr,sun2024coser}, style transfer \cite{zhang2023inversion,chung2024style}, text-based image editing \cite{brooks2023instructpix2pix,hertz2022prompt}, few-shot learning \cite{zhou2024advancing,giannone2022few,tan2023diffss}. Nevertheless, an inherent limitation of diffusion models lies in their poor controllability, which brings more challenges to fine-grained editing tasks, especially those that require interactive-style manipulation \cite{DragYourNoise,zhao2025real,zhou2025streaming}. 

Recent studies \cite{DragDiffusion,StableDrag} have explored the use of diffusion models for Drag-Based Image Editing (DBIE), which enables users to manipulate images by directly dragging objects via a set of user-specified handle and target points. Existing diffusion-based DBIE methods predominantly employ a point-based alternating optimization strategy \cite{DragYourNoise,StableDrag,FreeDrag,EasyDrag,DragText,DragDiffusion}, where \emph{Step-\textcolor{black}{1}}: optimizing the features of handle points toward corresponding target positions by performing point motion supervision; \emph{Step-\textcolor{black}{2}}: updating handle point positions iteratively via KNN-based point tracking.

\begin{figure}[t]
  \centering
   \includegraphics[width=1\linewidth]{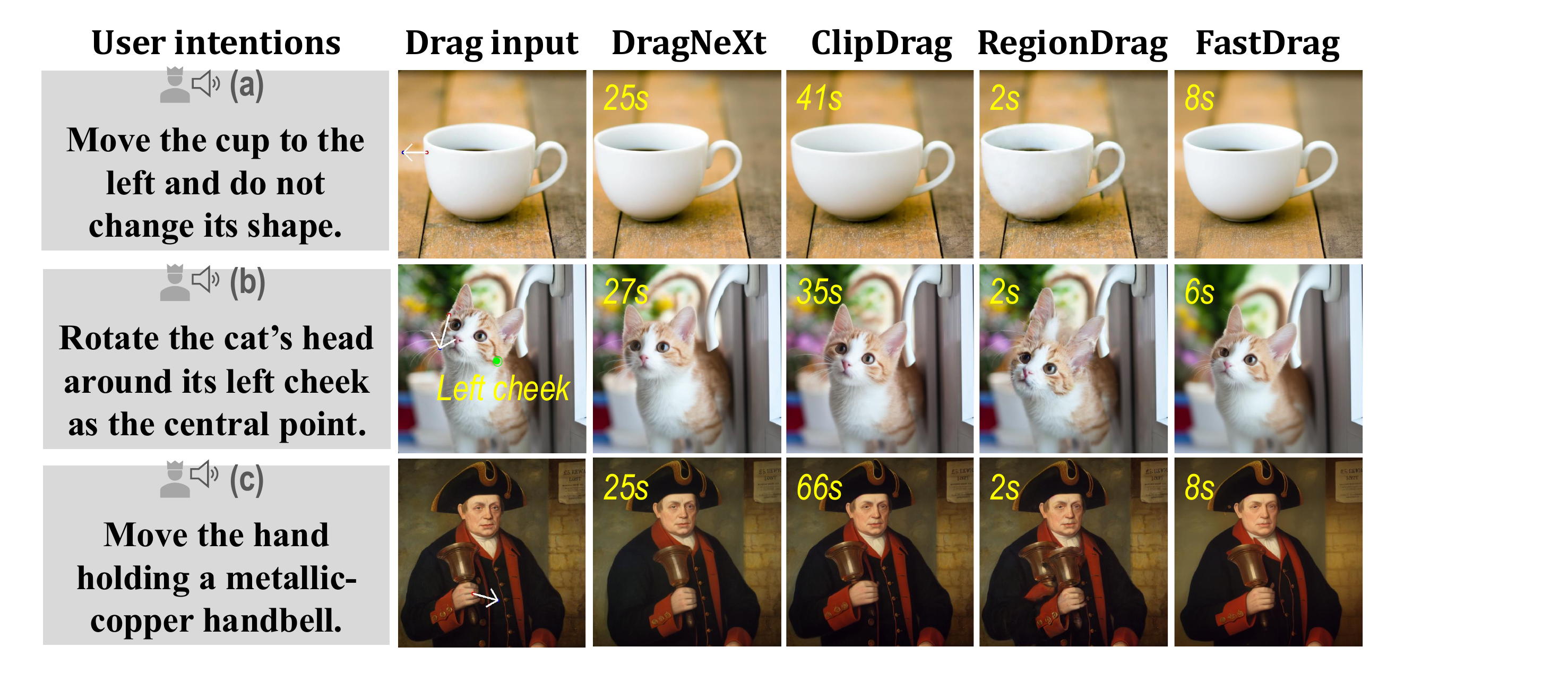}
   \caption{Examples of the key issues in current DBIE. (\emph{\textcolor{black}{i}}) Text prompts used in ClipDrag \cite{CLIPDrag} remain insufficient for solving the ambiguity issue; (\emph{\textcolor{black}{ii}}) predefined mapping functions employed by FastDrag \cite{FastDrag} and RegionDrag \cite{RegionDrag} boost efficiency but severely compromise editing quality. The numbers given in the upper-left corner of images indicate the latency for dragging the regions of handle points to target positions. 
   }
   \label{figure:1}
\end{figure}

However, the point-based alternating workflow inevitably brings two issues to DBIE: (\emph{\textcolor{black}{i}}) point-based drag suffers from high ambiguity and struggles to align with users' intentions, thereby severely compromising the precision of the editing process; (\emph{\textcolor{black}{ii}}) tackling DBIE through an alternating procedure of motion supervision and point tracking is not only cumbersome but also fails to always yield high-quality results, as accurately estimating the updated positions of handle points in each drag iteration is both challenging and time-consuming \cite{FreeDrag,StableDrag}. Besides, given that point-based motion supervision offers only limited structural cues about visual scenes, it cannot effectively guide DBIE.

Recently, ClipDrag \cite{CLIPDrag} sought to mitigate ambiguity by incorporating constraints from text prompts. Nonetheless, as a form of high-level descriptions, texts are often too vague to provide control signals required by fine-grained image manipulation \cite{DragDiffusion,ControlNet}. For example, as shown in Figure~\ref{figure:1}, even with the guidance of the prompt ``rotate the cat's head around its left cheek as the central point'', ClipDrag still fails to achieve the desired outcome. To boost DBIE's efficiency, FastDrag \cite{FastDrag} and RegionDrag \cite{RegionDrag} proposed using predefined mapping functions, rather than the learnable alternating paradigm. However, unfortunately, the warpage function and the copy-and-paste strategy used in FastDrag and RegionDrag are not flexible enough to handle all editing tasks and are prone to yielding unrealistic or unnatural results---such as the distorted cat’s head, deformed handbell, and visible artifacts in the edited areas shown in Figure \ref{figure:1}---thus severely degrading image quality.

Point‑based motion supervision and tracking are cumbersome and often difficult to align with users' intentions, whereas relying solely on the warpage function or the copy-and-paste strategy is far from delivering high-quality results. \textbf{These observations naturally lead us to ask two questions: \textcolor{black}{$\mapsto$}\textbf{$\bm{\mathcal{Q}}$1.} \emph{Is there a more effective solution to solve the ambiguity issue?} \textcolor{black}{$\mapsto$}\textbf{$\bm{\mathcal{Q}}$2.} \emph{How to enhance the efficiency of DBIE approaches based on alternating point motion supervision and tracking, while further improving their editing quality?}}

These two questions motivate us to revisit DBIE from a new perspective---unifying it as a Latent Region Optimization (LRO) problem, which aims to leverage region-level geometric transformations to optimize latent embeddings and realize drag manipulation. Therefore, by specifying the regions and types of geometric transformations, we can effectively address  \textbf{$\bm{\mathcal{Q}}$1}. Furthermore, we design a simple-yet-effective editing framework, \textcolor{SkyBlue}{\textbf{DragNeXt}}, to tackle \textbf{$\bm{\mathcal{Q}}$2}. For efficiency, it unifies DBIE as LRO and thus eliminates the necessity of conducting handle point tracking by upgrading point motion supervision to region-level optimization of latent embeddings. For editing quality, we propose a Progressive Backward Self-Intervention (PBSI) strategy that solves LRO by fully leveraging region-level self-intervention from intermediate drag states. By bypassing point tracking and considering region-level guidance from intermediate states, it can achieve a better trade-off between efficiency and quality.

\noindent\textbf{Contribution Summary}: (\emph{\textcolor{black}{i}}) We propose to unify DBIE as an LRO problem. Therefore, by specifying the regions and types of geometric transformations, we can effectively resolve the ambiguity issue. (\textcolor{black}{\emph{ii}}) We propose a simple yet effective editing framework, \textcolor{SkyBlue}{\textbf{DragNeXt}}, which tackles DBIE via LRO and further enhancing editing quality via performing PBSI. (\emph{\textcolor{black}{iii}}) We introduce NextBench, a benchmark with explicit user-intention annotations for evaluating alignment between user expectations and edited results. (\textcolor{black}{\emph{iv}}) On NextBench, the extensive experiments demonstrate that our \textcolor{SkyBlue}{\textbf{DragNeXt}} can achieve a better trade-off between editing efficiency and quality.

\section{Related Work}
DragDiffusion \cite{DragDiffusion} is the first work using diffusion models to achieve DBIE, which followed \cite{DragGAN} and conducted motion supervision and point tracking alternately. Based on \cite{DragDiffusion}, GoodDrag \cite{GoodDrag} further enhanced the fidelity of dragged areas by rearranging the drag process across multiple denoising timesteps. DragText \cite{DragText} proposed refining text embeddings to avoid drag halting. DragonDiffusion \cite{DragonDiffusion} and DiffEitor \cite{DiffEitor} discarded the tracking phase and directly applied point motion supervision between initial handle points and target points. To estimate handle point positions more accurately, StableDrag \cite{StableDrag} proposed a discriminative point tracking strategy, and FreeDrag \cite{FreeDrag} designed a line search backtracking mechanism. EasyDrag \cite{EasyDrag} advanced \cite{DragDiffusion} via introducing a stable motion supervision, which is beneficial for improving the quality of final results. FastDrag \cite{FastDrag} and RegionDrag \cite{RegionDrag} improved the efficiency of DBIE by employing fixed predefined mapping functions, where \cite{RegionDrag} is based on copy-and-paste and thus requires users to specify both handle and target areas. ClipDrag \cite{CLIPDrag} reduced the ambiguity of DBIE via using text prompts. DragNoise \cite{DragYourNoise} proposed editing on UNet's bottleneck features, which inherently contain more semantic information and can stabilize dragging. 

\noindent
\textbf{REMARK 1}. Our method  differs from DragDiffusion, GoodDrag, DragText, StableDrag, FreeDrag, EasyDrag, and DragNoise fundamentally, since it does not rely on alternating between point motion supervision and tracking. Instead of only considering initial relationships between handle and target points, as in DragonDiffusion and DiffEditor, we fully leverage progressive region-level guidance from intermediate drag states. CLIPDrag and RegionDrag overlook the ambiguity issue arising from the type of geometric transformations, whereas our approach does not rely on texts to reduce ambiguity. Different from FastDrag and RegionDrag, we only use geometric mapping functions to provide interventional signal and our learnable backward self-intervention strategy can fully leverage inherent prior knowledge learned by diffusion models via back-propagated gradients, avoiding unnatural deformation led by using a fixed transformation pattern.

\section{Methodology}
\label{methodology}
\subsection{Preliminaries}
\label{sec:prel}
\textbf{Diffusion Models}. Diffusion models \cite{ho2020denoising,rombach2022high,dhariwal2021diffusion} are composed of a diffusion process and a reverse process. During the diffusion, an image $\bm{x}$ is  encoded into latent space $\bm{z}_0$ and undergoes a gradual addition of Gaussian noise, $q(\bm{z}_t|\bm{z}_0)=\mathcal{N}(\sqrt{\alpha_t}\bm{z}_0,(1-\alpha_t)\bm{I})$, where $\alpha_t$ is a non-learnable parameter and decreases \emph{w.r.t.} the timestep $t$. The reverse process is to recover $\bm{z}_0$ from $\bm{z}_T$ by training a denoiser $\bm{\varepsilon}_{\bm{\Theta}}(\cdot)$:
\begin{equation}
    \mathcal{L}_{\bm{\Theta}} = \mathbb{E}_{t\sim\mathcal{U}(1,T), \bm{\varepsilon}_t\sim\mathcal{N}(0,I)} \left[ ||\bm{\varepsilon}_t - \bm{\varepsilon}_{\bm{\Theta}}(\bm{z}_t; t, \bm{c})||^2 \right]\
\label{eq:1}
\end{equation}
where $\bm{\varepsilon}_t$ denotes the groundtruth noise in the timestep $t$, and $\bm{c}$ represents an extra condition. Following the prior works \cite{DragYourNoise,DragDiffusion,GoodDrag}, we employ DDIM \cite{DDIM} in our approach.

\noindent\textbf{Drag-Based Image Editing}. Given $n$ pairs of handle and target points $\bm{\mathcal{O}}=\{\bm{h}_i=(x_i^h,y_i^h), \bm{g}_i=(x_i^g, y_i^g)\}_{i = 1,..., n}$, DBIE aims to edit an image $\bm{x}$ by dragging objects or regions indicated by handle points to target ones. Usually, an extra binary mask $\bm{M}$ is used to specify the uneditable region of $\bm{x}$.

\noindent\textbf{Motion Supervision and Point Tracking}. Current DBIE methods \cite{DragDiffusion,GoodDrag,DragText,StableDrag,FreeDrag,EasyDrag,DragYourNoise} mainly rely on performing motion supervision and point tracking alternately, where the former aims to transfer the features of handle points to target positions while the latter updates handle points iteratively and prevents dragging halt. We use $\mathcal{F}_{\bm{h}_i/\bm{g}_i}(\bm{z}_t)$ to denote the features extracted by $\bm{\varepsilon}_{\bm{\Theta}}(\cdot)$ at the location $\bm{h}_i$ or $\bm{g}_i$. Therefore, the objective function of the motion supervision can be described by Equation (\ref{eq:2}):
\begin{equation}
  \mathcal{L}_{m}(\bm{z}_{t}^{k}) = \sum_{i=1}^{n} \sum_{\bm{q} \in \bm{\pi }(\bm{h}_{i}^{k})} ||\mathcal{F}_{\bm{q}+\bm{d}_{i}}(\bm{z}_{t}^{k}) - \mathcal{SG}(F_{\bm{q}}(\bm{z}_{t}^{k}))||_{1} + \mathcal{R}_{\bm{M}}
\label{eq:2}
\end{equation}
where $\bm{z}_{t}^{k}$ and $\bm{h}^k_i$ denote the latent code $\bm{z}_{t}$ and the handle point $\bm{h}_i$ updated by $k$ iterations, $\bm{d}_i = (\bm{g}_i-\bm{h}^k_i)/||\bm{g}_i-\bm{h}^k_i||_2$ is the normalized vector from $\bm{h}^k_i$ to $\bm{g}_i$, $\bm{\pi}(\bm{h}_{i}^{k})$ denotes the neighborhood of $\bm{h}^k_i$, $\mathcal{SG}(\cdot)$ stops gradients from being back-propagated to variables, and $\mathcal{R}_{\bm{M}}$ is a constraint term to ensure the consistency of uneditable regions. After the motion supervision in each iteration $k$, point tracking is performed:
\begin{equation}
  \bm{h}_i^{k+1} = \mathop{\arg\min}_{\bm{q} \in \bm{\pi}(\bm{h}_i^k)} ||\mathcal{F}_{\bm{q}}(\bm{z}_t^{k+1}) - \mathcal{F}_{\bm{h}_i}(\bm{z}_t)||_1
\label{eq:3}    
\end{equation}
where $\mathcal{F}_{\bm{h}_i}(\bm{z}_t)$ indicates the features of the initial handle point $\bm{h}_i$ in the original latent code $\bm{z}_t$.

\noindent
\textbf{REMARK 2.} \textcolor{black}{$\mapsto$}\textbf{Why is point tracking critical for motion-based methods?} Point-based motion supervision is too local to provide enough guidance for the whole editing procedure. Losing the positions of handle points will severely interrupt the drag process, as no alternative guidance for editing remains, thereby significantly damaging the quality of edited images. \textcolor{black}{$\mapsto$}\textbf{What are limitations of methods based on motion supervision and point tracking?} Firstly, although the use of point tracking can alleviate the inherent limitation of point-based motion supervision, it is still very challenging to precisely estimate the updated positions of handle points. Inaccurate coordinate estimation can significantly mislead the dragging process, resulting in suboptimal outcomes.  Secondly, the alternating execution of Equation (\ref{eq:2}) and (\ref{eq:3}) results in low efficiency of DBIE, since motion supervision is repeatedly disrupted by iterative point tracking. Thirdly, point motion supervision suffers from high ambiguity and easily leads to gaps between user expectations and actual results. Last but not least, while using dense points can reduce ambiguity in some situations, this will substantially decrease the efficiency of the alternating workflow, as shown in Figure~\ref{figure:7}.

\subsection{Latent Region Optimization for Reliable Drag-Based Image Editing}
\label{sec:rdbie}
We begin by outlining key factors behind the ambiguity issue of DBIE in \textbf{Proposition \ref{pro:ambiguity}}, which can be further summarized as two key questions: \emph{how to drag?} and \emph{what to drag?}
\begin{proposition}[\textcolor{black}{\textbf{Key Factors to Ambiguity}}]
\label{pro:ambiguity}
\emph{The ambiguity of DBIE is twofold:} \textcolor{black}{$\mapsto$}\textbf{\emph{Factor-1}.} drag operations inherently involve multiple types---such as translation, deformation, and rotation---and treating them as type-agnostic induces ambiguity about users' intentions (\textbf{how to drag?}); \textcolor{black}{$\mapsto$}\textbf{\emph{Factor-2}.} point indicators are insufficient for accurately specifying objects or regions that need to be dragged (\textbf{what to drag?}). 
\end{proposition}
\begin{figure}[t]
\centering
\begin{minipage}[t]{0.495\linewidth} 
  \centering
  \includegraphics[width=0.95\linewidth]{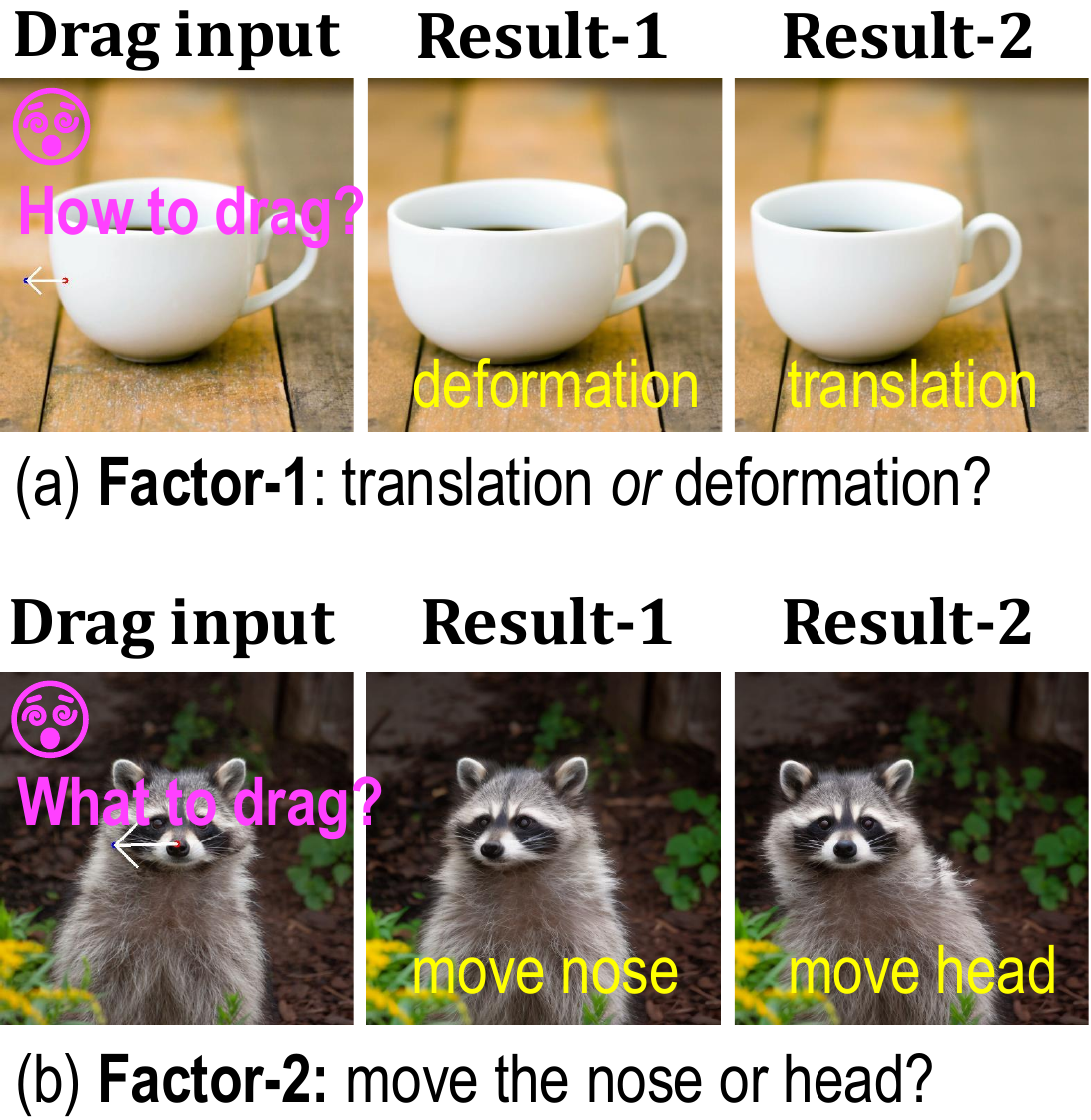}
  \caption{Factor-1 and -2.}
  \label{figure:2}
\end{minipage}%
\hfill
\begin{minipage}[t]{0.495\linewidth}
  \centering
  \includegraphics[width=0.95\linewidth]{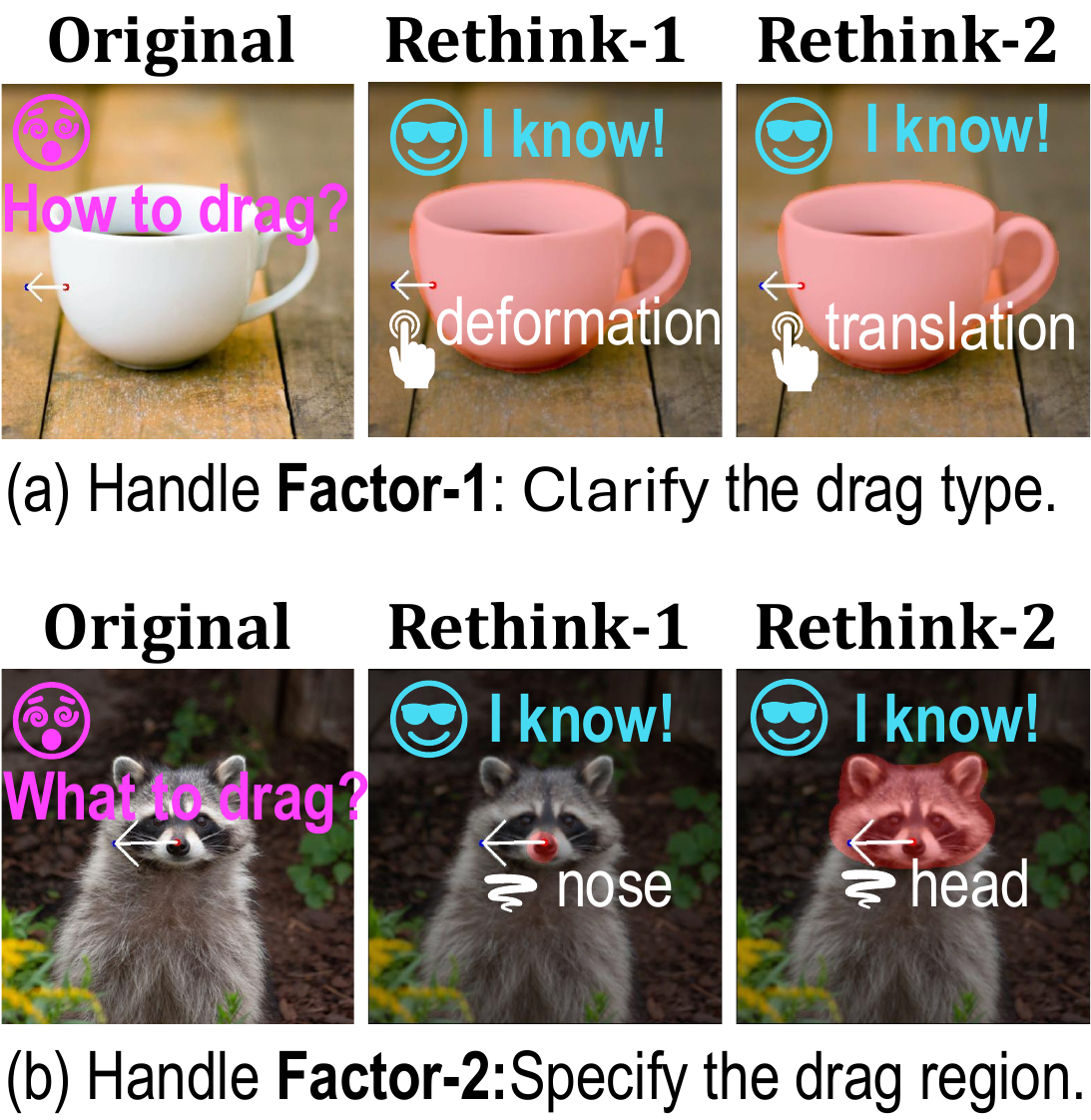}
  \caption{Rethink DBIE.}
  \label{figure:3}
\end{minipage}
\end{figure}

In Figure \ref{figure:2}, we provide an illustration for the two key factors, \textbf{Factor-1} and \textbf{Factor-2}. On one hand, the drag operation in Figure \ref{figure:2}(a) is inherently ambiguous since it could be interpreted as either a translational movement of the cup or a deformation of its edge region. This ambiguity stems from uncertainty about the types of drag operations (\emph{how to drag?}), which inevitably increases gaps between user expectations and model behaviors, thus damaging the precision of the editing process. On the other hand, in Figure \ref{figure:2} (b), the drag instruction could be either dragging the raccoon's nose, its head, or even its whole body. This type of ambiguity arises from uncertainty about which areas or objects to drag (\emph{what to drag?}) since points are too ambiguous to clearly reflect users' intentions. How to drag and what to drag are two fundamental problems in DBIE. Although textual description appears to be a shortcut, it actually does not work well as exemplified in Figure~\ref{figure:1}. We argue there is \textbf{No Free Lunch} in resolving these ambiguity issues, which means it is necessary to enable models to perceive drag operation types and areas in a more explicit way and design a more effective approach to guide them toward producing user-intended results. 

\noindent
\textbf{REMARK 3.} \textcolor{black}{$\mapsto$}Some previous methods have noticed the ambiguity issue in DBIE, but few of them consider both \textbf{Factor-1} and \textbf{Factor-2}, or provide a systematic and clear analysis for this problem, which we believe is critical and valuable for inspiring the further development of DBIE. 

\textbf{Based on the above observations, we step-by-step introduce how to explore DBIE from a new perspective, i.e., unifying DBIE as a Latent Region Optimization (LRO) problem.} We first rethink the DBIE task in \textbf{Proposition} \ref{pro:rethink}.
\begin{proposition}[\textcolor{gray!200}{\textbf{\textcolor{black}{Rethink DBIE}}}]
\label{pro:rethink}
DBIE can be regarded as performing geometric transformations on user‑specified regions of images.
\end{proposition}
\noindent
For instance, Result-1 given in Figure \ref{figure:2} (a) can be seens as applying a deformation transformation to the white coffee cup region. Form this perspective, by specifying the regions and types of geometric transformations, we can effectively resolve the ambiguity issue caused by Factor‑1 and Factor‑2, because of clarifying both how and what to drag as illustrated in Figure~\ref{figure:3}.

\begin{figure}[t]
  \centering
   \includegraphics[width=0.82\linewidth]{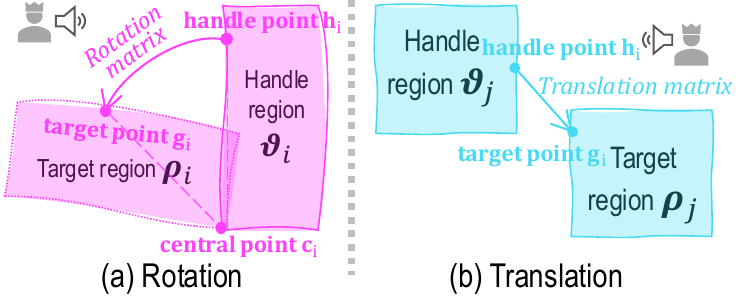}
   \caption{Examples of estimating target regions.} 
   \label{figure:a}
\end{figure}

Based on \textbf{Proposition \ref{pro:rethink}}, we further give the definition of our region-level \textbf{Reliable Drag-based Image Editing} (Reliable DBIE) in \textbf{Definition} \ref{def:dbie}, which aims to help users to yield reliable editing results and narrow gaps between their expectations and actual outcomes.
\begin{definition}[\textcolor{black}{\textbf{Reliable DBIE}}]
\label{def:dbie}
Reliable DBIE is to manipulate user-specified regions $\bm{\mathcal{E}}=\{\bm{\vartheta}_i\}_{i=1,...,n}$ of an image $\bm{x}$ based on the corresponding geometric transformations $\bm{\Gamma}=\{f_i\}_{i=1,...,n}$ inferred from instructions given by users.
\end{definition}

\begin{figure*}[t]
  \centering
   \includegraphics[width=0.76\linewidth]{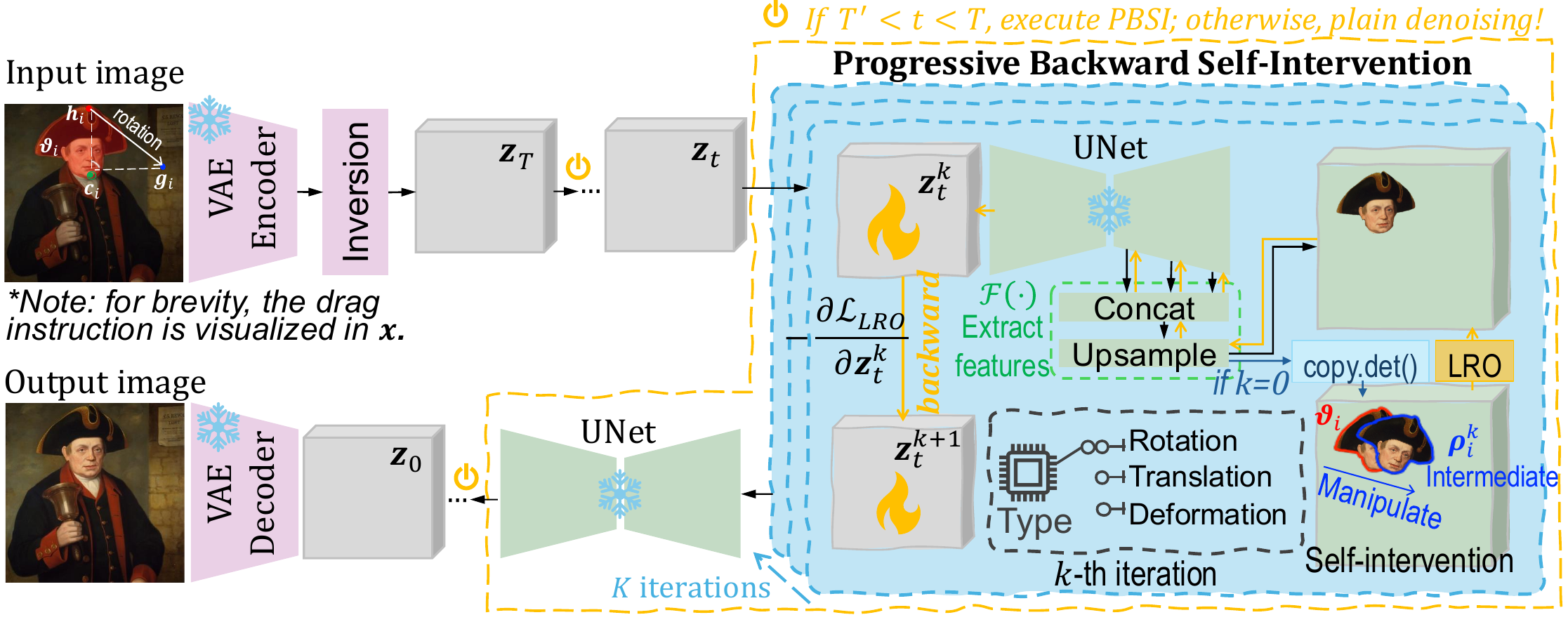}
   \caption{A brief illustration of our \textcolor{SkyBlue}{DragNeXt.}}
   \label{figure:4}
\end{figure*}

\noindent
Currently, the editing process is primarily performed on noise latent embeddings encoded by diffusion models, as they are more editable than original images \cite{Nulltext,Dreambooth}. Thus, we can further extend \textbf{Definition \ref{def:dbie}} to \textbf{Definition \ref{def:lro}}, and unify DBIE as a Latent Region Optimization (LRO) problem.  
\begin{definition}[\textcolor{black}{\textbf{\emph{Unify DBIE as LRO}}}]
\label{def:lro}
DBIE can be unified as optimizing specific target regions $\bm{\mathcal{P}}=\{\bm{\rho}_i\}_{i=1,...,n}$ within a latent code $\bm{z}_t$ based on user-specified handle regions $\bm{\mathcal{E}}=\{\bm{\vartheta}_i\}_{i=1,...,n}$ and the corresponding geometric transformations $\bm{\Gamma}=\{f_i\}_{i=1,...,n}$ involved in user instructions: 
 \begin{equation}
  \bm{z}^*_t = \mathop{\arg\min}_{\bm{z}_t}\mathcal{L}_{LRO}\left( \bm{z}_t,\{\bm{\rho}_i\}_{i=1,...,n}\right)
 \end{equation} 
 \begin{equation*}
 s.t., \{\bm{\rho}_i\}_{i=1,...,n}=\delta (\bm{\mathcal{E}},\bm{\Gamma})
 \end{equation*}
  where $\mathcal{L}_{LRO}$ is the objective function of LRO, and $\delta(\cdot)$ aims to produce binary masks $\{\bm{\rho}_i\}_{i=1,...,n}$ to identify target regions required to be optimized in $\bm{z}_t$ according to $\bm{\mathcal{E}}$ and~$\bm{\Gamma}$.
\end{definition}

\noindent
\textbf{REMARK 4.} 
\textcolor{black}{$\mapsto$}\textbf{Why LRO?} LRO serves as a bridge between DBIE and region‑level geometric transformations. Therefore, we can leverage many well‑studied geometric transformation functions in computer graphics to realize more reliable DBIE via explicitly controlling the dragging process. \textcolor{black}{$\mapsto$}\textbf{What can LRO do?} Different from  methods based on alternating point motion supervision and tracking, LRO takes into account region-level visual information, which provides more robust guidance for latent code manipulation. Under such regional supervision, it is unnecessary to excessively focus on positions of some specific points, as there exists sufficient context information to guide dragging. \textcolor{black}{$\mapsto$}\textbf{How to estimate target latent regions?} Target latent regions are estimated using geometric transformation functions widely adopted in computer graphics. As exemplified in Figure~\ref{figure:a}, if users intend to rotate the handle region $\bm{\vartheta}_i$, the region can be multiplied by a rotation matrix to achieve the desired geometric transformation.

\textcolor{black}{We observe that rotation and translation geometric transformations can cover most DBIE scenarios.} We reformulate DBIE's user input: users specify a set of handle regions $\bm{\mathcal{E}}=\{\bm{\vartheta}_i\}_{i=1,...,n}$ for an input image and give the corresponding drag instructions $\bm{\mathcal{C}}=\{\mathcal{T}_i,\bm{\mathcal{O}}_i\}_{i=1,...,n}$. If the operation type $\mathcal{T}_i=$``rotation'', $\bm{\mathcal{O}}_i=\{\bm{h}_i,\bm{g}_i,\bm{c}_i\}$ where $\bm{h}_i$ and $\bm{g}_i$ denote a pair of a handle point and a target point, and $\bm{c}_i$ represents a rotation center of $\bm{\vartheta}_i$; otherwise, $\bm{\mathcal{O}}_i=\{\bm{h}_i,\bm{g}_i\}$. Also, a binary mask $\bm{M}$ is adopted to specify the uneditable region. Based on $\{\bm{h}_i,\bm{g}_i\}$ or $\{\bm{h}_i,\bm{g}_i,\bm{c}_i\}$, the transformation function $f_i$ can be constructed by determining the corresponding rotation and translation matrix. \textcolor{black}{For details on converting input points to $f_i$, please refer to the supplementary material.}

\noindent
\textbf{\textcolor{black}{REMARK 5}. DBIE via regional geometric transformations.} \textbf{Object movement} can be achieved by translating an object' entire region; \textbf{deformation} can be realized by translating only its edge region; \textbf{2D rotation} can be achieved by applying a rotation transformation; and \textbf{3D rotation} can be interpreted as translating the sub-region of an object, assisted by prior knowledge inherently learned in
diffusion models (as shown in Figure \ref{figure:5} (j), the car’s 3D rotation can be realized by translating its front to the right). Explicit geometric functions and region-level guidance can help achieve better DBIE.

\subsection{Progressive Backward Self-Intervention: Less Meets More!}
\label{dragnext}
Based on \textbf{Definition \ref{def:lro}}, we further design \textcolor{SkyBlue}{\textbf{DragNeXt}} to enhance both editing quality and efficiency. As mentioned before, the alternating workflow lowers the efficiency of DBIE, while inaccurate handle point tracking easily leads to dragging halt and makes results unsatisfactory. Therefore, \textcolor{SkyBlue}{\textbf{DragNeXt}} addresses DBIE from an LRO perspective, eliminating the need for KNN‑based point tracking by explicitly advancing point-based motion supervision to region‑level optimization of latent embeddings. Moreover, it employs a Progressive Backward Self-Intervention (PBSI) strategy, which does not require accurately tracking point positions but still achieves superior editing results by fully leveraging progressive region-level guidance from intermediate transformation states.

\noindent\textbf{Progressive Backward Self-Intervention}. Figure \ref{figure:4} gives a brief illustration of our approach. Given an input image $\bm{x}$, we first encode it into latent space and perform DDIM inversion to produce $\bm{z}_{T}$. Then, PBSI is conducted from $T$ to $T'$ during denoising with $K$ iterations per timestep. We take the handle region $\bm{\vartheta}_i$ at the $k$-$th$ iteration of the timestep $t$ as an example to illustrate PBSI. We first extract the features of $\bm{z}_t^k$ by concatenating outputs from the last upsample blocks of all stages of $\bm{\varepsilon}_{\bm{\Theta}}(\bm{z}_t^k)$ and upsampling them to half of the resolution of $\bm{x}$, denoted as $\mathcal{F}(\bm{z}_t^k)$. Then, we estimate the intermediate transformation state $\bm{\rho}_{i}^{t,k}$ for the handle region $\bm{\vartheta}_i$ within the extracted features $\mathcal{F}(\bm{z}_t^k)$ based on user-given conditions $\bm{\mathcal{C}}$, which can be described by Equation (\ref{eq:5}):
\begin{equation}
\bm{\rho}_{i}^{t,k},\bm{\Pi}_{\bm{\vartheta}_i\rightarrow\bm{\rho}_{i}^{t,k}}=\delta (\bm{\vartheta}_i, f_i^{t,k})
\label{eq:5}
 \end{equation}
\begin{equation*}
s.t., \delta (\bm{\vartheta}_i, f_i^{t,k})=
\begin{cases}
  \texttt{Rot}(\bm{\vartheta}_i,\bm{c}_i,\theta), & \text{if $\mathcal{T}_i=$``rotation''} \\
  \texttt{Trans}(\bm{\vartheta}_i,\bm{\omega}), & \text{else}.
\end{cases}
\end{equation*}
In the equation, $\texttt{Rot}(\bm{\vartheta}_i,\bm{c}_i,\theta)$ aims to rotate the handle region $\bm{\vartheta}_i$ around the center point $\bm{c}_i$ by an angle $\theta=\eta^{t,k}*\angle\bm{h}_i\bm{c}_i\bm{g}_i$, $\texttt{Trans}(\bm{\vartheta}_i,\bm{\omega})$ translates $\bm{\vartheta}_i$ according to the offset vector $\bm{\omega}=\eta^{t,k}*(\bm{g}_i-\bm{h}_i)$, and $\eta^{t,k}=\frac{K*(T-t)+k}{K*(T-T'+1)}$ is a weighting factor that determines angles or offsets of intermediate states. Also, $\bm{\rho}_{i}^{t,k}$ is a binary mask that identifies the target intermediate region in $\mathcal{F}(\bm{z}_t^k)$, and $\bm{\Pi}_{\bm{\vartheta}_i\rightarrow\bm{\rho}_{i}^{t,k}}$ represents the coordinate mapping from the handle region $\bm{\vartheta}_i$ to the intermediate state $\bm{\rho}_{i}^{t,k}$.
Finally, we copy and detach the features extracted from the original latent code, $\mathcal{F}'(\bm{z}_t)=\mathcal{F}(\bm{z}_t)\texttt{.copy.detach()}$. Moreover, we interventionally adjust the detached features according to the obtained coordinate mapping, $\mathcal{F}'(\bm{z}_t)[\bm{\Pi}_{\bm{\vartheta}_i\rightarrow\bm{\rho}_{i}^{t,k}}]$, thereby perturbing the original latent representations and transforming the features of the handle region $\bm{\vartheta}_i$ to the intermediate target position $\bm{\rho}^{t,k}_i$. We consider self-intervention from the perturbed features to $\mathcal{F}(\bm{z}_t^k)$ and back-propagate interventional signal to the latent code $\bm{z}_t^k$ along the denoiser to update latent features. This can be depicted by Equation (\ref{eq:6}) and (\ref{eq:7}):
\begin{equation}
  \bm{z}_t^{k+1} \longleftarrow\bm{z}_t^k - \frac{\partial\mathcal{L}_{LRO}}{\partial\bm{z}_t^k},
\label{eq:6}
\end{equation}
\begin{equation}
\mathcal{L}_{LRO}=\|\mathcal{F}(\bm{z}_t^k)*\bm{\rho}_{i}^{t,k}-\mathcal{F}'(\bm{z}_t)[\bm{\Pi}_{\bm{\vartheta}_i\rightarrow\bm{\rho}_{i}^{t,k}}]*\bm{\rho}_{i}^{t,k}\|_1+\mathcal{R}_{\bm{M}}.
\label{eq:7}
\end{equation}
Minimizing $\mathcal{L}_{LRO}$ back-propagates self-intervention gradients to the latent code, thus progressively dragging handle regions to target positions. Once PBSI is complete, we denoise $\bm{z}_{T'}$ to $\bm{z}_{0}$ and decode it into image space. The pseudocode of our \textbf{\textcolor{SkyBlue}{DragNeXt}} is provided in Algorithm 1 of the appendix.

\begin{figure*}[t!]
\centering
\includegraphics[width=0.85\linewidth]{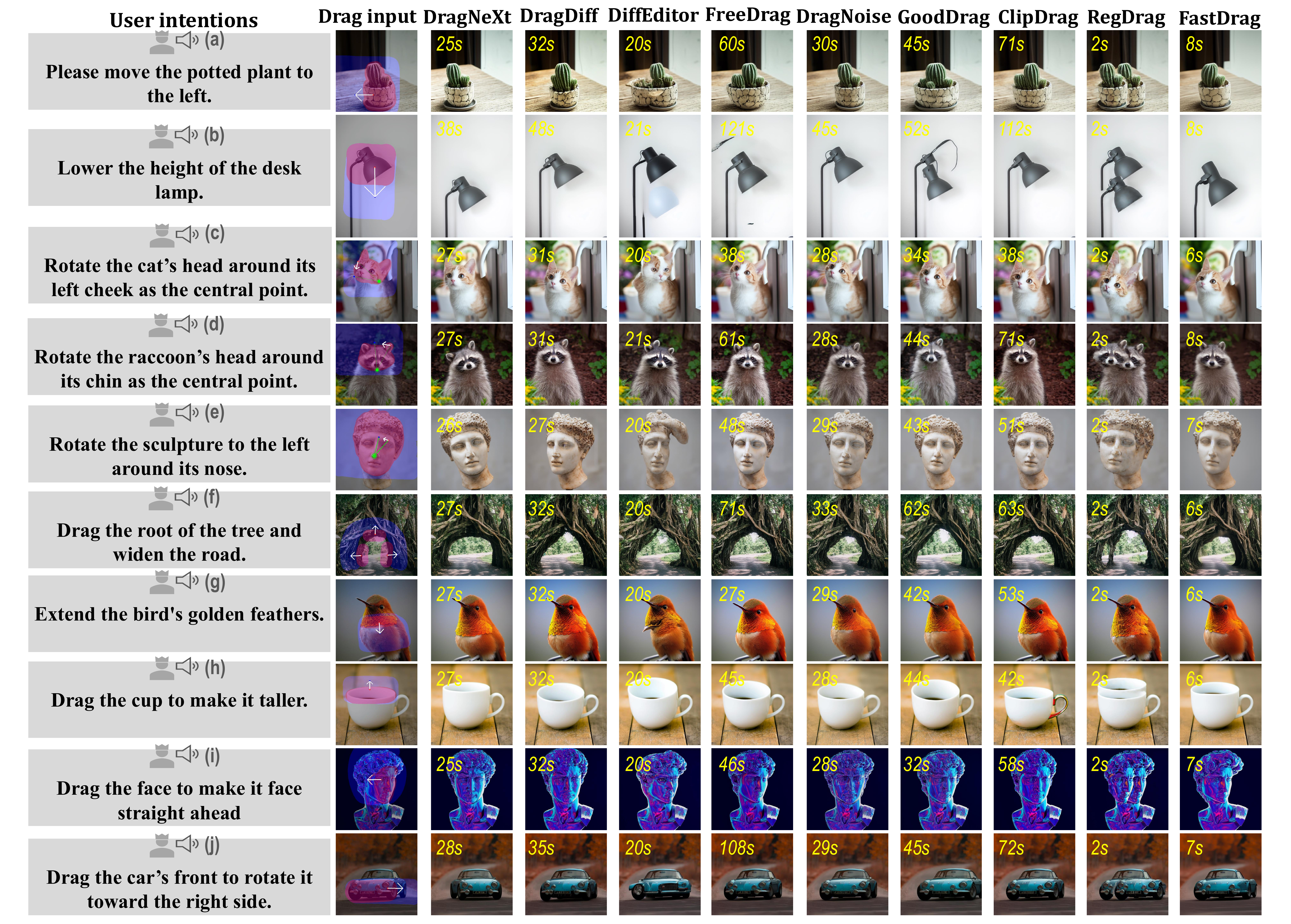}
\caption{Qualitative results achieved by our \textcolor{SkyBlue}{DragNeXt}.}
\label{figure:5}
\end{figure*}

\noindent
\textbf{REMARK 6}. \textcolor{black}{$\mapsto$}\textbf{Why backward self-intervention?} Our method also adopts geometric mapping functions. However, unlike RegionDrag and FastDrag, which directly use them to manipulate latent code, we instead leverage them to provide interventional signal. By optimizing latent code through back-propagated gradients from the denoiser, our approach fully exploits the prior of pretrained diffusion models, thereby mitigating unnatural results caused by fixed mapping functions. \textcolor{black}{$\mapsto$}\textbf{Difference between Equation (\ref{eq:7}) and (\ref{eq:2}).} $\mathcal{L}_{LRO}$ considers region-level guidance, whereas $\mathcal{L}_{m}$ performs point supervision and needs to iteratively track handle point positions. \textcolor{black}{$\mapsto$}\textbf{\textcolor{black}{Discussion on Equation (\ref{eq:5}}}). We unify translation, deformation, and 3D rotation into a single mapping function, $\texttt{Trans}(\cdot)$, as deformation and 3D rotation can be interpreted by translating the partial regions of an object, assisted by priors inherently encoded in diffusion models, \emph{e.g.,} in Figure~\ref{figure:5}~(j), the car’s 3D rotation can be achieved by translating its front to the right. We extend drag points to regional guidance to more clearly specify regions to move, deform, or rotate.

\begin{figure}[t]
    \centering
     \includegraphics[width=0.9\linewidth]{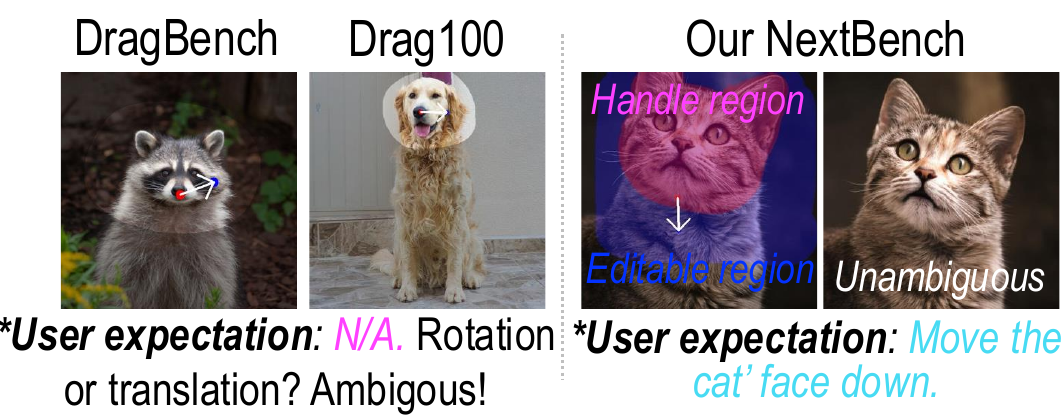}
     \caption{Comparison between our NextBench and the previous benchmarks, DragBench and Drag100.} 
     \label{figure:b}
\end{figure}

\section{Experiments}
\label{experiment}
We first introduce our NextBench and evaluation metrics, followed by the main results of our method and ablation studies. We provide implementation details in the supplementary material due to the limited space of the paper's main body.

\subsection{NextBench: a Benchmark for Reliable DBIE}
To better evaluate model performance on Reliable DBIE, we propose a new benchmark, \textbf{NextBench}, that comprises 234 test samples with drag operations including translation, 2D/3D rotation, and deformation. Each sample is clearly annotated with user intentions to better assess how well model outputs align with user expectations, as shown in Figure \ref{figure:b}.

\noindent
\textbf{Why NextBench?} Existing benchmarks, such as DragBench \cite{DragDiffusion} and Drag100 \cite{GoodDrag}, still contain ambiguous drag instructions, \emph{e.g.}, as shown in Figure~\ref{figure:b}, the raccoon’s and dog’s heads could either be translated or rotated, and both would be considered as satisfactory results. Our NextBench explicitly annotates each sample with a clear user expectation, handle regions, and editable areas, therefore enabling a more reliable assessment of intention–result alignment and regional consistency. NextBench also treats 3D and 2D rotations as two distinct operations, showing that current approaches struggle with 2D rotation and excel at yielding 3D rotation based on the prior of pretrained diffusion models.

\subsection{Evaluation Metrics}
Following prior work \cite{GoodDrag,DragDiffusion}, we use LPIPS and DAI to evaluate performance on NextBench, where the radius of DAI is set as 20. To better assess region-level DBIE, LPIPS is computed between original images and edited results in three parts: (\emph{\textcolor{black}{i}}) LPIPS$_{ue}$ for uneditable regions (\emph{a lower LPIPS$_{ue}$ indicates better preservation of uneditable regions}); (\emph{\textcolor{black}{ii}}) LPIPS$_{th}$ for consistency between handle and target regions (\emph{a lower LPIPS$_{th}$ means handle regions are successfully dragged to target positions}); (\emph{\textcolor{black}{iii}}) LPIPS$_{hh}$ for handle regions (\emph{Successful dragging handle regions to target positions should result in a higher $\text{LPIPS}_{hh}$ between the handle regions of original input images and edited results, reflecting the change of visual content}).


\begin{table}[t]
        \centering
        \small
        \resizebox{0.47\textwidth}{!}{
        \begin{tabular}{c|c|c|c|c|c}
            \hline
           \textbf{Method} & \textbf{Lat}$\downarrow$  & \textbf{DAI$\downarrow$} & \textbf{LPIPS$_{ue}\downarrow$} & \textbf{LPIPS$_{th}\downarrow$} & \textbf{LPIPS$_{hh}\uparrow$}  \\
            \hline
            DragDiff & 36s  &  0.10333 & 0.05914 & 0.32314 & 0.23663 \\
            DiffEditor & 24s  &  0.07255 &   0.05960 & 0.30570 &  0.23291\\
            DragNoise & 34s  & 0.10001 & 0.05891 &  0.35697 & 0.24789 \\
            FreeDrag & 69s  & 0.11113  &  0.07693 &  0.36907 & 0.26723  \\
            GoodDrag & 51s  & 0.06878 & \textbf{0.05794} & 0.28667 & 0.25291 \\
           ClipDrag & 58s &  0.11909 & 0.06243 & 0.37292 & 0.20528 \\
            RegionDrag & \textbf{3s}  & \underline{0.05968} & 0.05930 &  \underline{0.18780} & 0.19736 \\  
            FastDrag & 8s  & 0.08530 & 0.06624 &  0.32646 & \underline{0.28411} \\
            \hline
            \textbf{\textcolor{SkyBlue}{DragNeXt}} & $28$s  & \textbf{0.05775} & \underline{0.05886} & \textbf{0.16223} & \textbf{0.36653} \\
            \hline
        \end{tabular}}
        \caption{Quantitative results on NextBench. ``Lat'' indicates mean latency per image for dragging handle point regions to target positions. $\uparrow / \downarrow$ denotes higher$/$lower values are better.}
        \label{tab:1}
\end{table}

\begin{figure*}[t]
  \centering
  \begin{minipage}[t]{0.21\textwidth}
    \centering
    \includegraphics[width=0.8\linewidth]{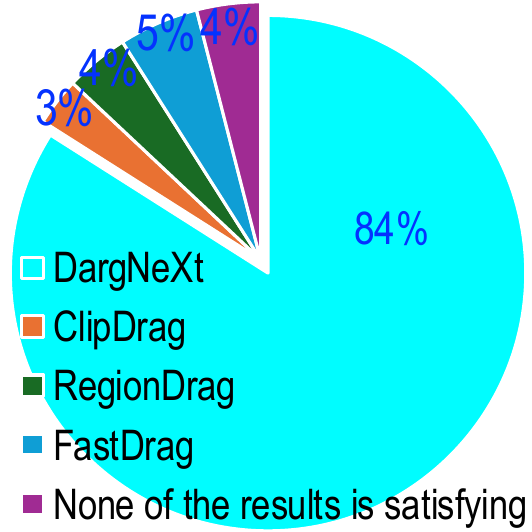}
    \caption{Voting results.}
    \label{figure:c}
  \end{minipage}
  \hfill
  \begin{minipage}[t]{0.78\textwidth}
    \centering
    \includegraphics[width=1\linewidth]{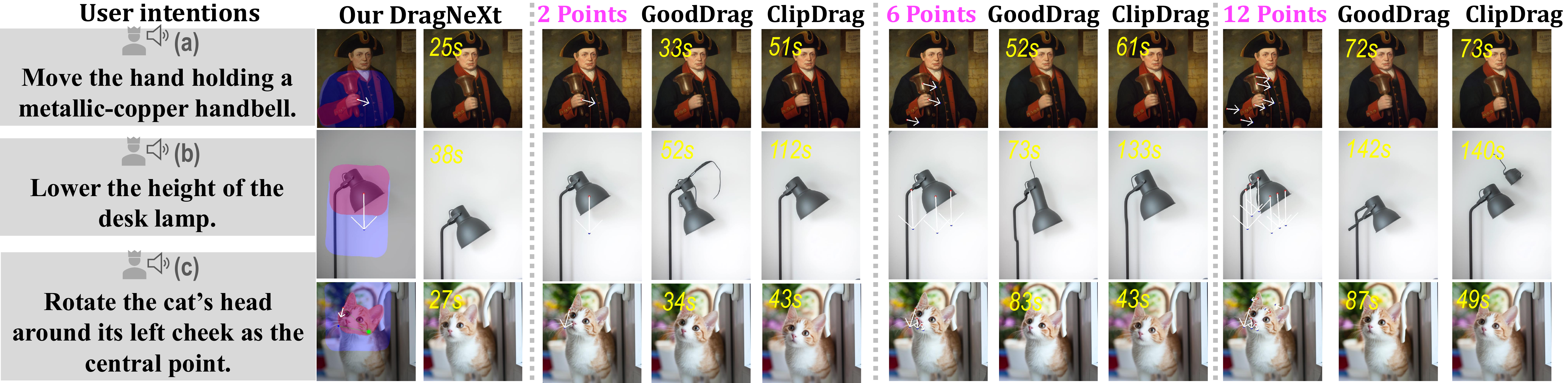}
    \caption{Efficiency and quality improvements over the point-based alternating workflow.}
    \label{figure:7}
  \end{minipage}
\end{figure*}

\subsection{Main results}
We compare our method with eight typical open-source DBIE methods: DragDiffusion, DiffEditor, DragNoise, FreeDrag, GoodDrag, ClipDrag, RegionDrag, and FastDrag.

\noindent
\textbf{Qualitative Results.} We present qualitative results of our method in Figure \ref{figure:5}, from which we can make the following observations. By specifying the regions and the types of geometric transformations, \textbf{\textcolor{SkyBlue}{DragNeXt}} achieves better alignment between user expectations and results, \emph{e.g.}, as shown in Figure \ref{figure:5} (a), our models translate the potted plant leftward without damaging its shape. Based on LRO and PBSI, \textbf{\textcolor{SkyBlue}{DragNeXt}} achieves a better trade-off between efficiency and quality, \emph{e.g.}, it has obviously higher efficiency than methods based on alternating point motion supervision and tracking,  while delivering obviously higher editing quality than those relying solely on predefined mapping functions.

\noindent
\textbf{REMARK 7.} \textbf{Discussion on 2D$/$3D rotation.} \textcolor{black}{$\mapsto$}We observe that existing DBIE methods particularly excel at yielding 3D rotation effects, \emph{e.g.}, as shown in Figure \ref{figure:5} (i) and (j), dragging the face or the vehicle front enables most methods to rotate them. This capability arises from the strong prior of pretrained diffusion models, which are creative to generate rotated objects based on latent features perturbed by drag operations. \emph{Among these methods, \textbf{\emph{\textcolor{SkyBlue}{DragNeXt}}} leverages region-level visual cues, which can obviously guide diffusion models to realize better 3D rotation effects}. \textcolor{black}{$\mapsto$}Current DBIE methods generally fail to perform 2D rotation, as these patterns are not well captured and learned by diffusion models. \textbf{\textcolor{SkyBlue}{DragNeXt}} can mitigate this by explicitly using regional geometric transformations.

\noindent\textbf{Quantitative Results.} The quantitative results are summarized in Table \ref{tab:1}. The table shows that our method achieves the lowest DAI and LPIPS$_{th}$ and the highest LPIPS$_{hh}$, demonstrating that our method can effectively drag objects from handle regions to target positions. Unsuccessfully dragging objects to target positions results in low LPIPS$_{hh}$---indicating little change in handle regions---and high LPIPS$_{th}$ due to mismatch between original handle regions and edited target regions. Also, the value of LPIPS$_{ue}$ indicates that our method can preserve the high fidelity of uneditable areas.

\noindent\textbf{Anonymous User Study}. We further conducted user studies to validate our method, where 26 participants are invited. Selecting the most appropriate result from too many options is time-consuming; to reduce volunteers’ workload, we include the three most relevant methods: ClipDrag, RegionDrag, and FastDrag. The results consistently demonstrate the superiority of \textbf{\textcolor{SkyBlue}{DragNeXt}}, \emph{e.g.,} 84\% of the votes favored our results, demonstrating higher quality and better alignment with user expectations (see the supplementary material for details).

\subsection{Method Analysis}
\noindent\textbf{Efficiency and Quality Improvements}. In Figure \ref{figure:7}, we analyze the efficiency and quality improvements of our method over the point-based alternating workflow. Without losing generality, we choose two recent typical works---GoodDrag and ClipDrag---as compared baselines. The results consistently validate the effectiveness of our method. The point-based methods are often difficult to align with user intentions, e.g., as shown in Figure \ref{figure:7} (a), GoodDrag and ClipDrag drag only the hand and leave the handbell unmoved using 2 points. Although increasing the number of points can mitigate ambiguity, this largely slows down the alternating drag process and fail to always guide models to yield satisfactory results.

\begin{figure}[t]
  \centering
   \includegraphics[width=1\linewidth]{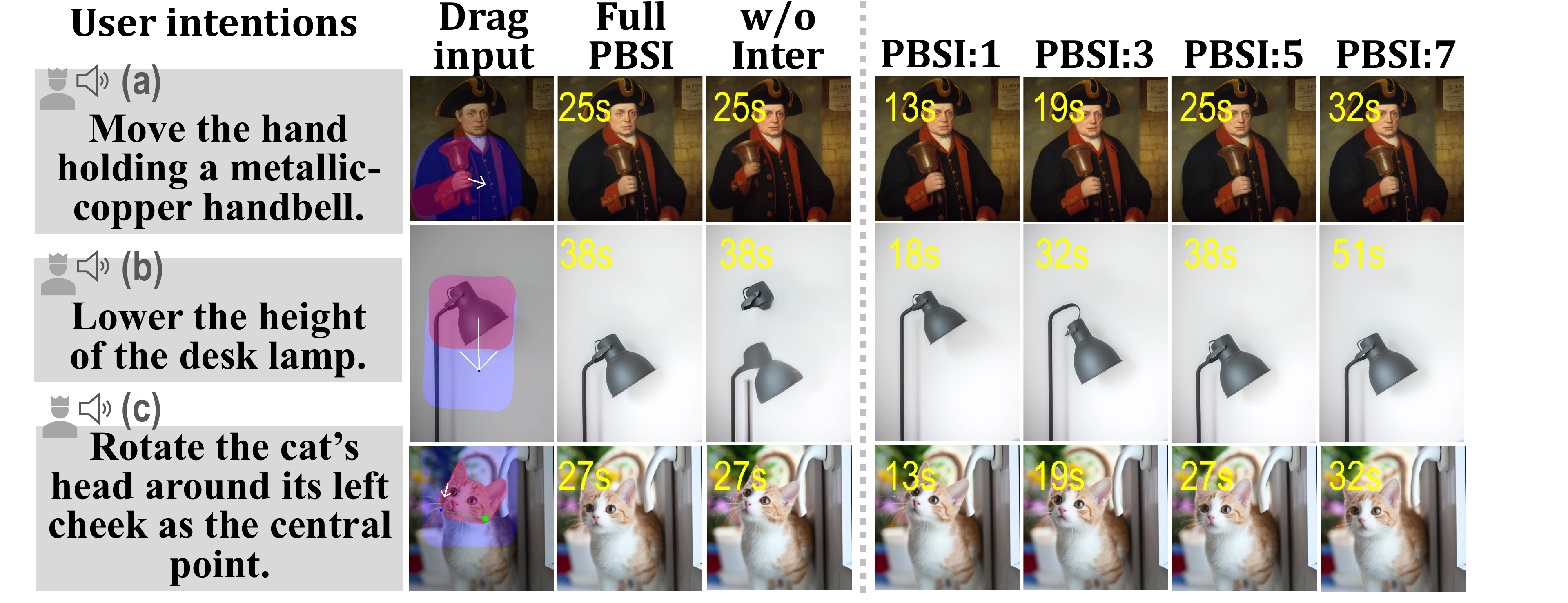}
   \caption{Ablation study on PBSI. ``Full PBSI'' indicates using the full PBSI strategy, ``w/o inter'' represents the guidance from intermediate drag states is not considered in PBSI, and ``PBSI: $N$'' indicates that PBSI is performed over $N$ timesteps. Zoom in for a better view.} 
   \label{figure:6}
\end{figure}

\noindent\textbf{Ablation Study on PBSI.} We provide ablation studies for our PBSI strategy in Figure \ref{figure:6}. Based on the results shown in the figure, we have the following observations. Firstly, removing the guidance of intermediate states significantly degrades output quality, \emph{e.g.}, the hand in Figure \ref{figure:6} (a) is dragged to an incorrect position, and unnatural results are yielded in Figure \ref{figure:6} (b), thereby demonstrating its important role in achieving high-quality results of DBIE. We also study the impact of performing PBSI over different numbers of timesteps. When PBSI is applied to only a single denoising timestep, objects cannot be successfully dragged to target positions. By contrast, increasing the timesteps of performing PBSI obviously improves the quality of edited results, saturating after $5$ timesteps, which also indicate the effectiveness of our method in guiding diffusion models to achieve DBIE.

\section{Conclusion}
\label{conclusion}
We propose to address Drag-Based Image Editing (DBIE) from a new perspective---unifying it as a Latent Region Optimization (LRO) problem that aims to use region-level geometric transformations to optimize latent code to realize drag-based manipulation. By specifying the areas and types of geometric transformations, we can effectively reduce gaps between users' intentions and actual model behaviors. We also design a new simple-yet-effective editing framework, dubbed \textcolor{SkyBlue}{\textbf{DragNeXt}}. It solves LRO through a Progressive Backward Self-Intervention (PBSI), which simplifies the procedure of DBIE while further enhancing editing quality by fully leveraging region-level structure information and progressive guidance from intermediate transformation states. Physically driven DBIE remains highly challenging. Therefore, in the future, we plan to enhance our \textcolor{SkyBlue}{\textbf{DragNeXt}} by integrating physics-based geometric transformation functions.

\section*{Acknowledgments}
This research is supported by the RIE2025 Industry Alignment Fund – Industry Collaboration Projects (IAF-ICP)
(Award I2301E0026), administered by A*STAR, as well as supported by Alibaba Group and NTU Singapore through Alibaba-NTU Global e-Sustainability CorpLab (ANGEL).

\bibliography{ref}


\section{A. DDIM Sampling and Inversion}
In this section, we provide more details about DDIM \cite{DDIM}, which is employed in our editing framework. DDIM defines the sampling of diffusion models as a non-Markovian process: 

\begin{align}
\label{eq:8}
&q(\bm{z}_{t-1}|\bm{z}_t,\bm{z}_0)=\\&\mathcal{N}\left(
\sqrt{\alpha_{t-1}} \bm{z}_0 + \sqrt{1 - \alpha_{t-1} - \sigma_t^2} \cdot 
\frac{\bm{z}_t - \sqrt{\alpha}_t \bm{z}_0}{\sqrt{1 - \alpha}_t},\  \alpha_t^2 \bm{I}
\right).  
\nonumber
\end{align} 
Therefore, it can be formulated by using Equation~(\ref{eq:9}):
\begin{align}
\label{eq:9}
\bm{z}_{t-1}&=
\sqrt{\alpha_{t-1}} \left( \frac{\bm{z}_t - \sqrt{1 - \alpha_t} \, \bm{\varepsilon}_{\bm{\Theta}}(\bm{z}_t)}{\sqrt{\alpha_t}} \right) \\&+
\sqrt{1 - \alpha_{t-1} - \sigma_t^2} \cdot \bm{\varepsilon}_{\bm{\Theta}}(\bm{z}_t) + \sigma_t \bm{\varepsilon},
\nonumber
\end{align}
where $\bm{\varepsilon}\sim\mathcal{N}(\bm{0},\bm{I})$ represents standard Gaussian noise and is independent of the latent code $\bm{z}_t$, and $\sigma_t = \eta \sqrt{(1 - \alpha_{t-1}) / (1 - \alpha_t)} \sqrt{1 - \alpha_t / \alpha_{t-1}}$ for all timesteps. When setting $\eta = 1$, Equation~(\ref{eq:9}) becomes DDPM, equaling to a stochastic differential equation (SDE). Setting $\eta = 0$ yields a deterministic sampling process, corresponding to an ordinary differential equation (ODE). Given the sampling process in Equation~(\ref{eq:9}), DDIM inversion can be described by Equation~(\ref{eq:10}):
\begin{align}
\label{eq:10}
\boldsymbol{z}_{t+1} = &\frac{\sqrt{\alpha_{t+1}}}{\sqrt{\alpha_{t}}} \left( \boldsymbol{z}_{t} - \sqrt{1 - \alpha_{t }} \cdot \bm{\varepsilon}_{\bm{\Theta}}(\bm{z}_t) \right) \\&+ \sqrt{1 - \alpha_{t+1}} \cdot\bm{\varepsilon}_{\bm{\Theta}}(\bm{z}_t),
\end{align}
which is based on the assumption that the ODE is invertible in the limit of small step sizes.

\section{B. Implementation Details}
\noindent
We implement our \textbf{\textcolor{SkyBlue}{DragNeXt}} in PyTorch and, following prior works~\cite{GoodDrag,RegionDrag,FastDrag,CLIPDrag}, employ \texttt{Stable-Diffusion-v1-5} as the base model to ensure fair comparison between methods. We optimize the learnable parameters using the Adam optimizer \cite{adam} with a learning rate of $2\times10^{-2}$. Following~\cite{DragDiffusion,GoodDrag}, we finetune diffusion models via LoRA~\cite{lora} with a rank of $16$. The number of denoising timesteps is set to $T_{\text{max}} = 50$, and the inversion strength is fixed at $0.75$, meaning that each input image is inverted to the timestep $T = 50\times0.75 = 38$, respectively. Also, $T'$ and $K$ are set to $33$ and $10$. Following~\cite{DragDiffusion,GoodDrag,RegionDrag}, we incorporate mutual self-attention~\cite{masactrl} starting from the $10$-th layer of UNet.

\section{C. NextBench: a Benchmark for Reliable DBIE}
\label{supp:nextbench}
\begin{figure*}[t!]
\centering
\includegraphics[width=1\linewidth]{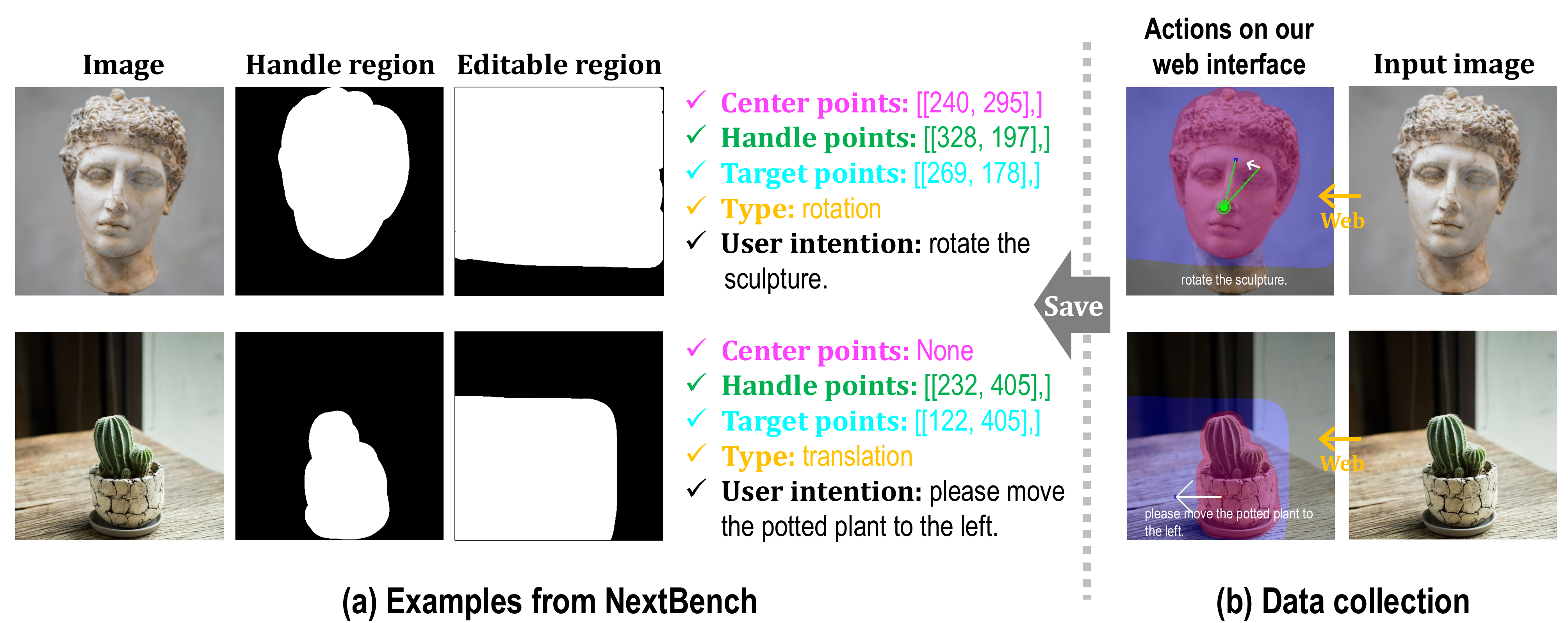}
\caption{\textbf{A brief illustration of samples from our NextBench.}} 
\label{figure:8}
\end{figure*}

To better assess model performance on Reliable DBIE, we propose a new benchmark, \textbf{NextBench}, comprising $234$ carefully annotated samples with detailed drag instructions and corresponding user intentions. As illustrated in Figure~\ref{figure:8} (a), each drag instruction is specified by six key components: handle regions, editable regions, center points, handle points, target points, and transformation types. Annotators explicitly record their intentions for each sample, enabling a more faithful evaluation of how well generated results align with user expectations.

NextBench is the first benchmark to explicitly incorporate constraints on both the type and region of geometric transformations for dragging, serving as a critical step toward realizing Reliable DBIE. To streamline data collection, we developed a user-friendly web-based system, following the pipeline illustrated in Figure~\ref{figure:8} (b), which will be publicly released soon. NextBench offers diverse content, including $200$ real images and $34$ AI-generated images, covering $103$ animal images, $18$ artistic paintings, $32$ landscapes, $24$ plant images, $31$ human portraits, and $26$ everyday objects such as furniture and vehicles. As a high-quality benchmark is essential for driving progress in this field, we are committed to the continuous maintenance and improvement of NextBench.

\noindent\textbf{Evaluation metrics.} Following prior work \cite{GoodDrag,DragDiffusion}, we use LPIPS and DAI to evaluate performance on NextBench. DAI aims to measure distance between patches centered at handle points and those centered at corresponding target points, which is first introduced by GoodDrag \cite{GoodDrag}:
\begin{equation}
\text{DAI} = \frac{1}{n} \sum_{i=1}^{n} 
\frac{\left\lVert \bm{\phi}(\bm{z}_0)_{\pi(\bm{h}_i;\gamma)} - \phi(\hat{\bm{z}}_0)_{\bm{\pi}(\bm{g}_i;\gamma)} \right\rVert_2^2}
{(1 + 2\gamma)^2},
\label{eq:dai}
\end{equation}
where $\phi$ denotes the VAE decoder that maps $\bm{z}_0$ to the RGB image space, and $\bm{\pi}(\bm{h}_i,\gamma)$ represents a patch centered at $\bm{h}_i$ with a radius $\gamma$. Equation (\ref{eq:dai}) computes the mean squared error between the patch at the handle point $\bm{h}_i$ of $\phi(\bm{z}_0)$ and the corresponding patch at the target point $\bm{g}_i$ of $\phi(\hat{\bm{z}}_0)$.

To better assess region-level DBIE, LPIPS is computed between original images and edited results in three parts: (\emph{\textcolor{black}{i}}) LPIPS$_{ue}$ for uneditable regions, (\emph{\textcolor{black}{ii}}) LPIPS$_{th}$ for consistency between handle and target regions, and (\emph{\textcolor{black}{iii}}) LPIPS$_{hh}$ for handle regions, which are shown in Equation (\ref{eq:11}), (\ref{eq:12}), and (\ref{eq:13}), respectively:
\begin{equation}
\label{eq:11}
\text{LPIPS}_{ue} = \frac{1}{m} \sum_{j=1}^{m} \texttt{LPIPS}(\bm{x}_j[\bm{M}], \bar{\bm{x}}_j[\bm{M}])
\end{equation}
\begin{equation}
\label{eq:13}
\text{LPIPS}_{th} = \frac{1}{mn} \sum_{j=1}^{m}\sum^{n}_{i=1} \texttt{LPIPS}(\bm{x}_j[\bm{\vartheta}_i], \bar{\bm{x}}_j[\bm{\rho}_i])
\end{equation}
\begin{equation}
\label{eq:12}
\text{LPIPS}_{hh} = \frac{1}{mn} \sum_{j=1}^{m}\sum^{n}_{i=1} \texttt{LPIPS}(\bm{x}_j[\bm{\vartheta}_i], \bar{\bm{x}}_j[\bm{\vartheta}_i]).
\end{equation}
In the above equations, $\texttt{LPIPS}(\cdot)$ measures LPIPS values between input images, and $[\cdot]$ selects regions where given binary masks have a value of $1$. $\bm{x}_j$ and $\bar{\bm{x}}_j$ represent a pair of an original image and an edited result, $\{\bm{\vartheta}_i\}_{i=1,...,n}$ represent handle regions given by users, and $\{\bm{\rho}_i\}_{i=1,...,n}$ denote target regions in edited results. Target regions can be estimated by considering drag instructions given by users, as mentioned in \textbf{Definition \ref{def:lro}} and Equation (\ref{eq:5}) of the paper's main body.

\noindent\textbf{What can these metrics do?} According to Equation (\ref{eq:11}), (\ref{eq:13}), (\ref{eq:12}), we summarize the functions of these metrics here:
\begin{itemize}
    \item LPIPS$_{ue}$ aims to measure LPIPS between the uneditable regions of an original input image and an edited result. A lower LPIPS$_{ue}$ indicates better preservation of uneditable regions, whereas a higher LPIPS$_{ue}$ implies that uneditable regions are altered during dragging.
    \item LPIPS$_{th}$ aims to measure the consistency between the handle regions of an original input image and the target regions of an edited result. A lower LPIPS$_{th}$ means handle regions are successfully dragged to target positions; vice versa, failing to drag objects to target positions results in a higher LPIPS$_{th}$ due to the mismatch between original handle regions and target areas in editable results.
    \item LPIPS$_{hh}$ is to measure divergence between the handle regions of an input image and an edited result. Successful dragging handle regions to target positions should result in a higher LPIPS$_{hh}$, reflecting the change of visual content; otherwise, handle regions in the original and edited image are the same, resulting in a lower LPIPS$_{hh}$.
\end{itemize}

\noindent\textbf{Why not point-based metrics?} We do not employ point-based evaluation metrics in NextBench, such as the Mean Distance (MD) between handle and target points \cite{DragDiffusion}, as they are incompatible with the region-based nature of our proposed Reliable DBIE. Unlike the previous point-based DBIE setting, Reliable DBIE emphasizes region-level consistency, rendering point-based metrics biased and insufficient for evaluating model performance in this context.

\section{D. Translation, Deformation, and 2D/3D Rotation}
\label{supp:Trans_Defor_Rotation}
In this work, we adopt two geometric transformations widely used in computer graphics: \textbf{translation} and \textbf{rotation}. We observe that these two transformation functions can cover most DBIE scenarios, including translation, deformation, and 2D/3D rotation. \textbf{Theoretically, our approach does not have restrictions on the types of geometric transformations and is compatible with other transformations used in computer graphics.} This paper aims to provide a new foundational framework for DBIE, and we leave the exploration of more transformations within this framework to our future research.

\subsection{Translation}
Translation refers to moving an object or region from one location to another withou  altering its shape or size. Suppose a point $\bm{p}=(x, y)$ is translated along a direction $\bm{d}=(d_x, d_y)$; its new coordinates $\bm{p}’=(x’, y’)$ can be computed using Equation~(\ref{eq:14}):
\begin{equation}
\label{eq:14}
\begin{bmatrix}
x' \\
y' \\
1
\end{bmatrix}
=
\underbrace{
\begin{bmatrix}
1 & 0 & d_x \\
0 & 1 & d_y \\
0 & 0 & 1
\end{bmatrix}
}_{\text{Translation matrix}}
\begin{bmatrix}
x \\
y \\
1
\end{bmatrix}
\end{equation}
where the first term on the right-hand side of the equation is commonly referred to as the translation matrix. Region-level translation is achieved by applying this translation to every point within the region. According to the handle point $\bm{h}_i$ and the target point $\bm{g}_i$, we can calculate the translation offset $\bm{d}_i$ about the region $\bm{\vartheta}_i$ as $\bm{d}_i={\bm{g}_i-\bm{h}_i}=(x^g_i-x^h_i, y^g_i-y^h_i)$. 

\begin{figure}[t!]
\centering
\includegraphics[width=1\linewidth]{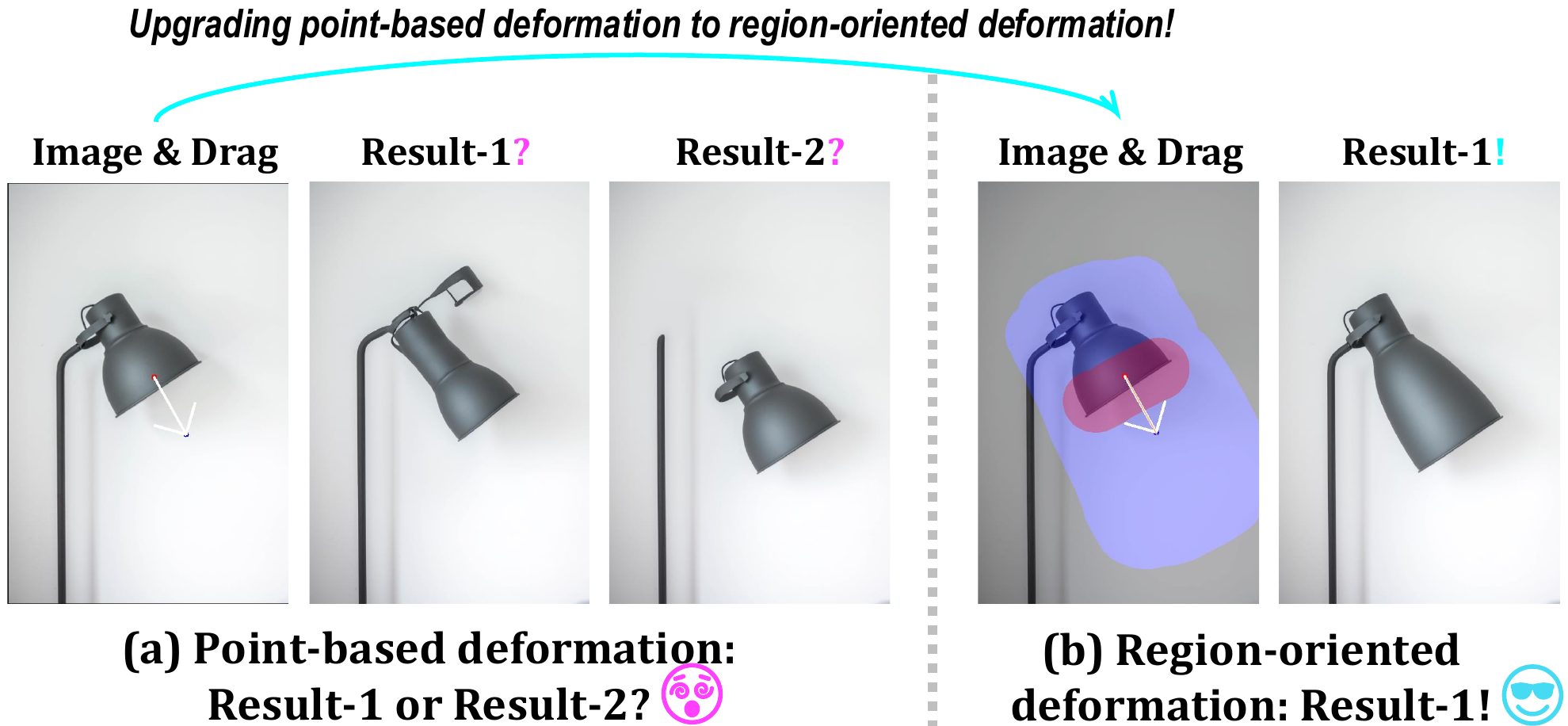}
\caption{\textbf{Point-Based VS. Region-Oriented Deformation.}}
\label{figure:suppa}
\end{figure}

\begin{figure}[h!]
\centering
\includegraphics[width=1\linewidth]{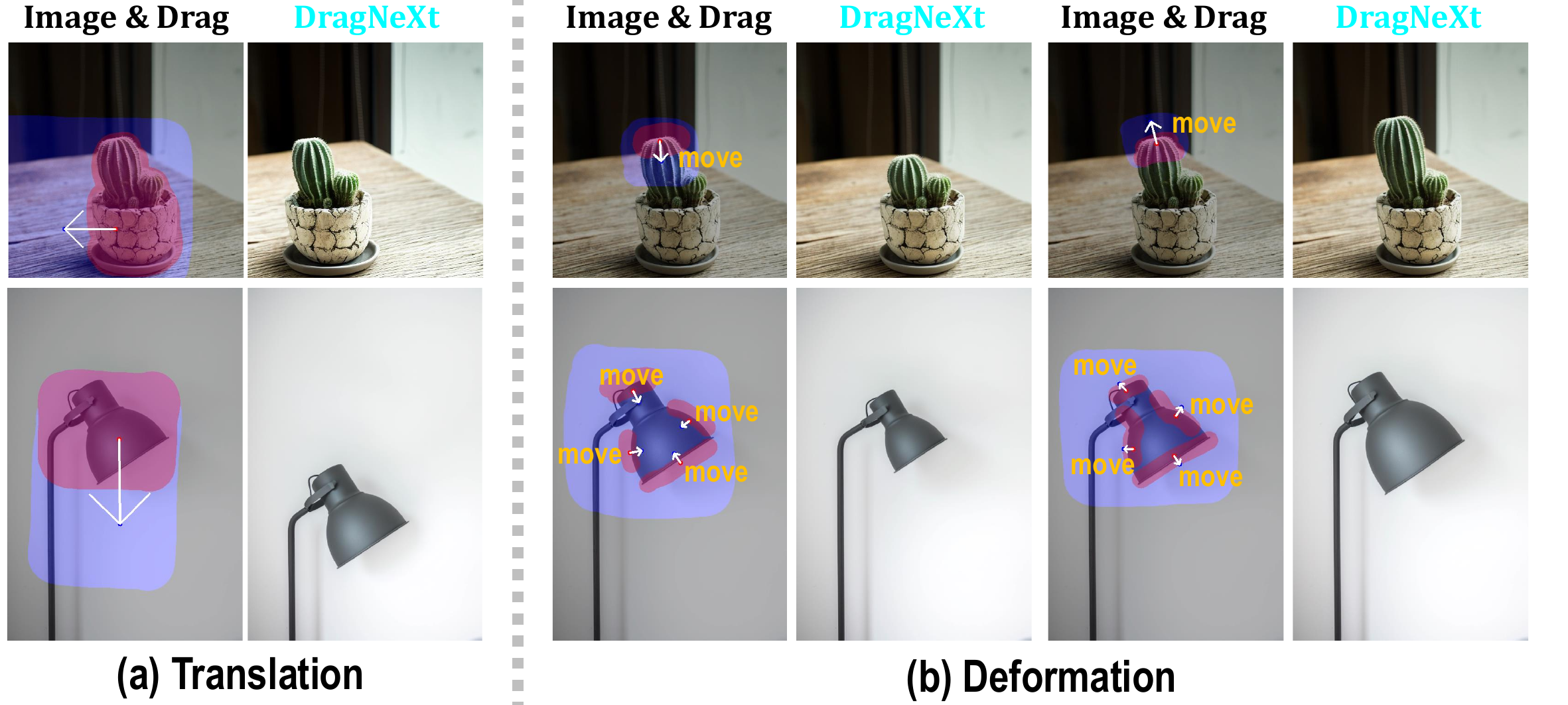}
\caption{\textbf{Object Translation VS. Deformation.} Translation is moving the entire region of an object, whereas deformation can be seen as translating an object's subregion.}
\label{figure:9}
\end{figure}
\begin{figure}[h!]
\centering
\includegraphics[width=1\linewidth]{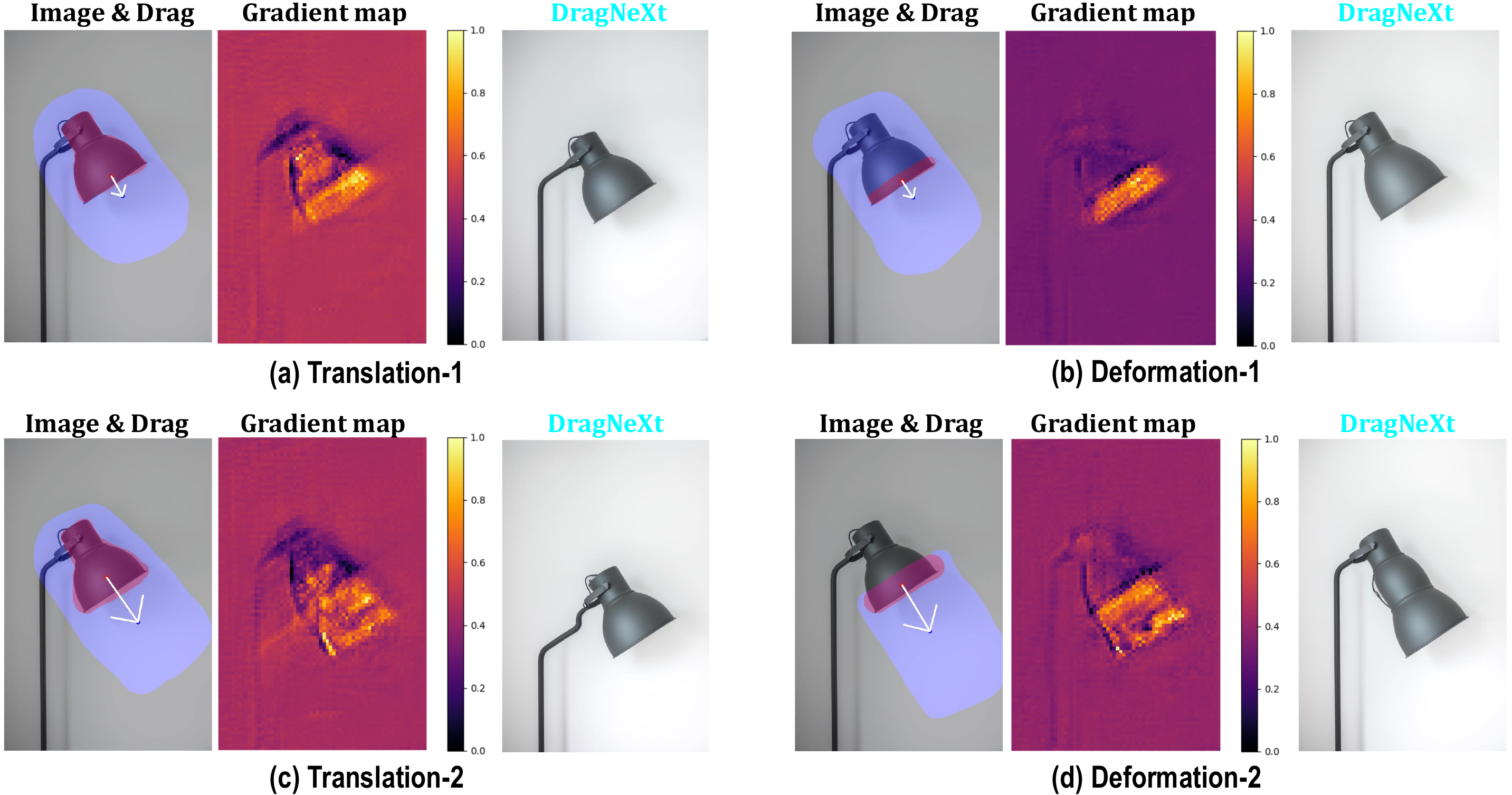}
\caption{\textbf{Visualization of gradients back-propagated to latent code} when dragging an object or its local area.} 
\label{figure:10}
\end{figure}
\subsection{Deformation}
Deformation refers to a non-rigid transformation that alters the shape of an object. As shown in Figure \ref{figure:suppa} (a), previous DBIE methods achieve object deformation by dragging handle points to corresponding target positions. In contrast, our \textbf{DragNeXt} extends this paradigm from point-based to region-oriented deformation by reformulating object deformation as the translation of the local part of an object, \emph{e.g.,} as shown in Figure \ref{figure:suppa} (b), dragging the edge region of the lamp to realize its deformation. In the following, we provide more discussions and analysis for our region-oriented deformation.

\noindent
\textbf{Point-Based VS. Region-Oriented Deformation.} Using points along to instruct the dragging process has a major drawback: \textbf{handle points can only offer limited contextual information and cannot accurately specify which regions should be dragged or deformed.} For example, the drag points in Figure \ref{figure:suppa} (a) can be interpreted as either dragging the whole lamp or the lamp's edge, introducing severe ambiguity into the dragging process. By contrast, extending point-based instructions to region-level guidance enables full exploitation of the information of pixels around handle points and can more clearly specify which regions need to be dragged. We also observe that incorporating region-level visual contexts leads to higher-quality deformation results, \emph{e.g.}, Result-1 in Figure \ref{figure:suppa} (b) is obviously better than Result-1 in Figure \ref{figure:suppa} (a). These observations validate the importance of extending point-based deformation to region-based deformation!

\noindent
\textbf{Region-Oriented Deformation VS. Translation.}
As can be seen from Figure \ref{figure:9}, object translation can be regarded as moving the entire region of an object, whereas deformation can be seen as moving an object' subregion. In this work, we extend the previous point-based deformation to region-oriented deformation, enabling more reliable DBIE by leveraging region-level visual context. However, similar to prior methods \cite{GoodDrag,CLIPDrag,DragonDiffusion,RegionDrag,FastDrag}, this approach still cannot achieve physics-driven deformation results. Currently, achieving physics-driven DBIE results remains highly challenging. We leave it for our future work, and incorporate physics-based transformation functions in our editing framework.

\noindent
\textbf{REMARK 8.} In Figure \ref{figure:10}, we provide a visualization of gradients back-propagated to latent code when dragging an object or its subregion. In Figure \ref{figure:10} (a) and (b), the object and its local region are dragged over a short distance. The gradients are primarily localized in the regions that require manipulation, whereas the areas that do not need adjustment remain unaffected. In Figure~\ref{figure:10} (c) and (d), the desk lamp and its local area are dragged over a longer distance. We observe that, regardless of the dragging distance, regions with distinct appearances remain separated from the target regions. For instance, the black desk lamp does not disturb the appearance of the white background in Figure \ref{figure:10} (c); in Figure \ref{figure:10} (d), the background does not affect the extended region of the desk lamp caused by the dragging operation. This phenomenon can be attributed to strong prior knowledge and patterns learned by pretrained diffusion models from vast amounts of training data.

\subsection{2D/3D Rotation}
Rotation refers to the process of rotating an object or region around a specified point by a certain angle. The rotation operation can be categorized into two types: 2D rotation and 3D rotation, which are one-by-one introduced below.

\begin{figure*}[h!]
\centering
\includegraphics[width=0.9\linewidth]{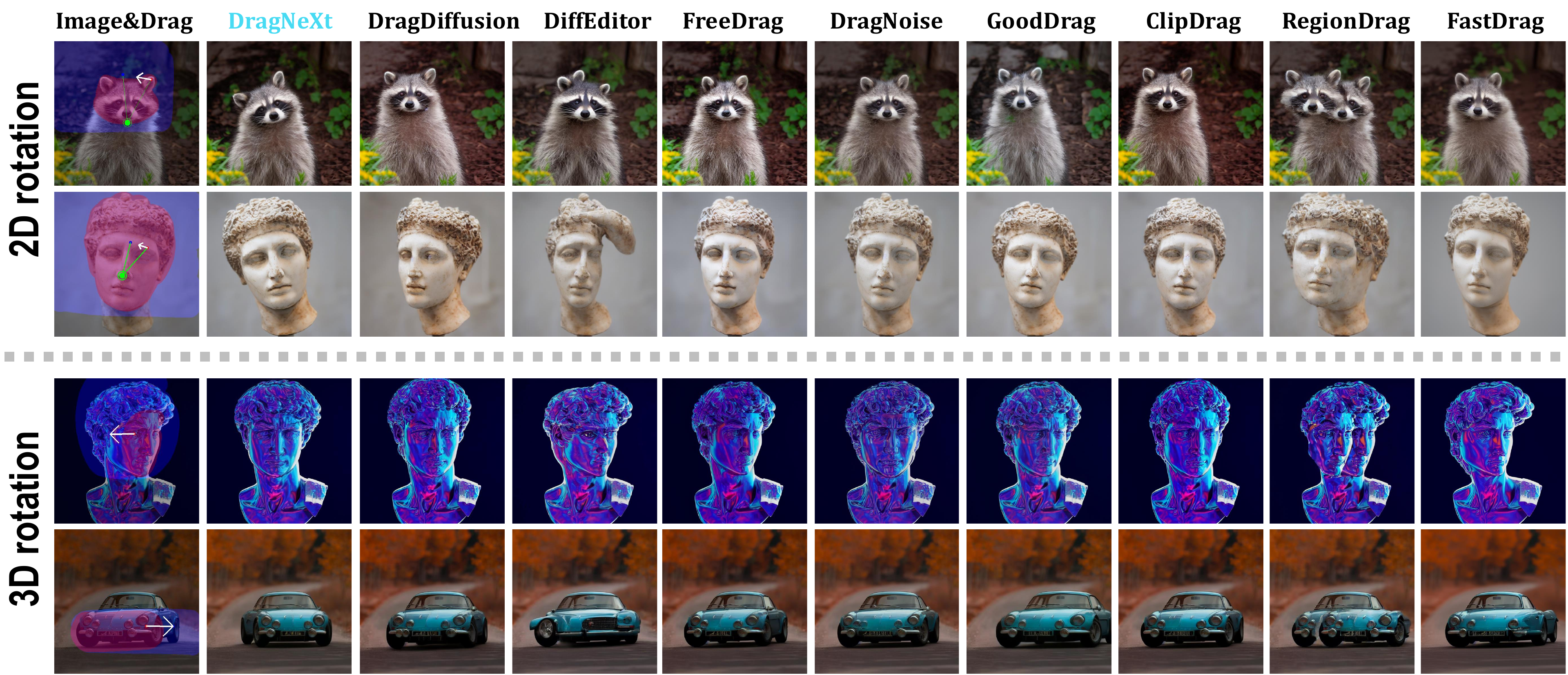}
\caption{\textbf{Illustrations of 2D Rotation and 3D Rotation.}} 
\label{figure:11}
\end{figure*}

\begin{figure}[h!]
\centering
\includegraphics[width=0.8\linewidth]{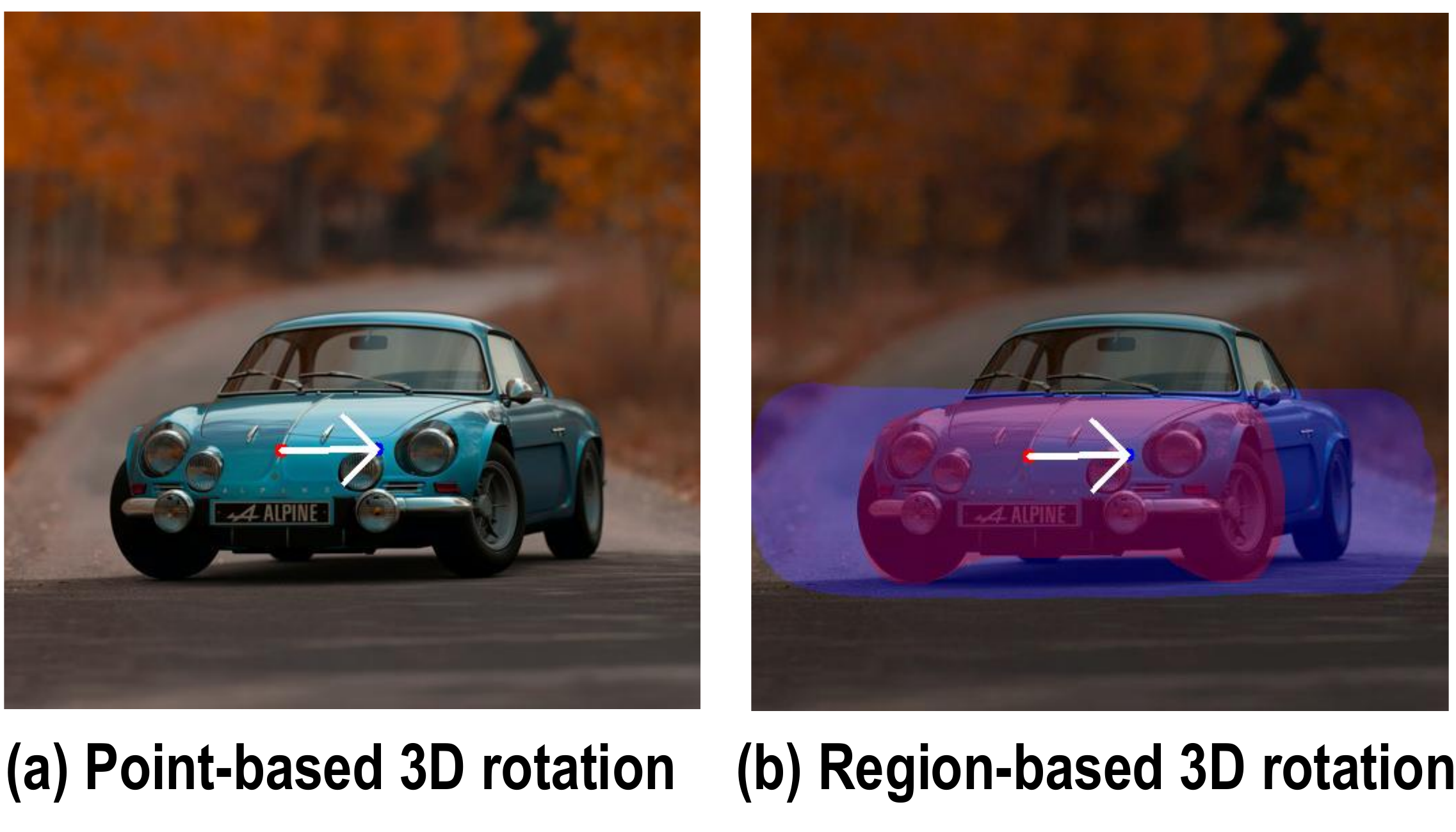}
\caption{\textbf{Point-Based VS. Region-Based 3D Rotation.} Point-based 3D rotation is introduced by GoodDrag \cite{GoodDrag}, and region-based 3D rotation is used in our \textbf{DragNeXt}. The figure (a) is borrowed from GoodDrag.} 
\label{figure:suppc}
\end{figure}

\noindent
\textbf{2D Rotation.} Suppose we aim to rotate a region around a center point $\bm{c} = (x_c, y_c)$ by an angle $\theta$. For each point $\bm{p} = (x, y)$ in this region, the updated coordinates $\bm{p}’ = (x’, y’)$ can be computed using Equation~(\ref{eq:15}):
\begin{align}
\label{eq:15}
\left[\begin{smallmatrix}
x' \\
y' \\
1
\end{smallmatrix}\right]
=
\underbrace{
\left[\begin{smallmatrix}
1 & 0 & x_c \\
0 & 1 & y_c \\
0 & 0 & 1
\end{smallmatrix}\right]
}_{\text{Back to }\bm{c}}
\underbrace{
\left[\begin{smallmatrix}
\cos\theta & -\sin\theta & 0 \\
\sin\theta & \cos\theta & 0 \\
0 & 0 & 1
\end{smallmatrix}\right]
}_{\text{Rotation matrix}}
\underbrace{
\left[\begin{smallmatrix}
1 & 0 & -x_c \\
0 & 1 & -y_c \\
0 & 0 & 1
\end{smallmatrix}\right]
}_{\text{To the origin}}
\left[\begin{smallmatrix}
x \\
y \\
1
\end{smallmatrix}\right],
\end{align}
where the middle term on the right-hand side of the equation is commonly referred to as the rotation matrix, while the remaining matrices are used to translate regions either to the origin or back to the center point $\bm{c}$. According to the handle point $\bm{h}_i=(x^h_i,y^h_i,)$, target point $\bm{g}_i=(x^g_i,y^g_i,)$, and center point $\bm{c}_i=(x^c_i,y^c_i,)$ given by users, we can calculate the rotation matrix as follows:  
{\scriptsize
\begin{equation}
\cos\theta =
\frac{(x_i^h - x_i^c)(x_i^g - x_i^c) + (y_i^h - y_i^c)(y_i^g - y_i^c)}
{\sqrt{(x_i^h - x_i^c)^2 + (y_i^h - y_i^c)^2} \;
 \sqrt{(x_i^g - x_i^c)^2 + (y_i^g - y_i^c)^2}}
\end{equation}
}
{\scriptsize
\begin{equation}
\sin\theta =
\frac{(x_i^h - x_i^c)(y_i^g - y_i^c) - (y_i^h - y_i^c)(x_i^g - x_i^c)}
{\sqrt{(x_i^h - x_i^c)^2 + (y_i^h - y_i^c)^2} \;
 \sqrt{(x_i^g - x_i^c)^2 + (y_i^g - y_i^c)^2}}.
\end{equation}
}

\noindent
\textbf{3D Rotation.} We compare 2D and 3D rotations in Figure~\ref{figure:11}. The 2D rotation operation can be explicitly modeled by using a geometric transformation function. However, there is no predefined mapping function capable of well handling 3D rotation, since it involves a complex non-rigid transformation that inherently depends on strong priors---such as object appearance, shape, and structural consistency---during the dragging process. For example, as shown in the last column of Figure~\ref{figure:11}, the warpage function, used in FastDrag \cite{FastDrag}, easily leads to unnatural and unrealistic deterioration of objects.

\noindent
\textbf{Observation 1. We have observed an interesting phenomenon in our experiments: \emph{despite 3D rotation being inherently more challenging than 2D rotation, current DBIE models surprisingly perform better on 3D rotation than on 2D rotation!}} 

\noindent
As shown in the third row of Figure~\ref{figure:11}, most DBIE methods are able to rotate the 3D angle of the sculpture by directly dragging the face leftward; by contrast, however, none of the methods can achieve 2D rotation of the face of the sculpture or the raccoon, as exhibited in the first and second rows.

\noindent
\textbf{We conclude the reasons for this counterintuitive phenomenon in twofold:} (i) the strong capability of current DBIE methods in 3D rotation actually arises from the strong prior of pretrained diffusion models, which are creative to generate rotated objects based on latent features perturbed by drag operations; (ii) 2D rotation can be explicitly modeled by using geometric transformation functions, but the 2D rotation pattern is not well captured and learned by diffusion models during pretraining.

\noindent 
\textbf{Point-Based Rotation VS. Region-Based Rotation.} The concept of 3D rotation in DBIE was formally introduced for the first time in GoodDrag \cite{GoodDrag}. As shown in Figure \ref{figure:suppc} (a), GoodDrag uses handle and target points to instruct the dragging process of 3D rotation, \emph{i.e.}, dragging handle points to the positions of target points. As shown in Figure \ref{figure:suppc} (b), we also extend the point-based 3D rotation to the region-based 3D rotation. Extending point-based instructions to region-level guidance enables
full exploitation of the information of pixels around handle
points and can more clearly specify which regions need to be
dragged, thereby helping models to achieve better 3D rotation effects.

\begin{figure*}[t!]
\centering
\includegraphics[width=0.75\linewidth]{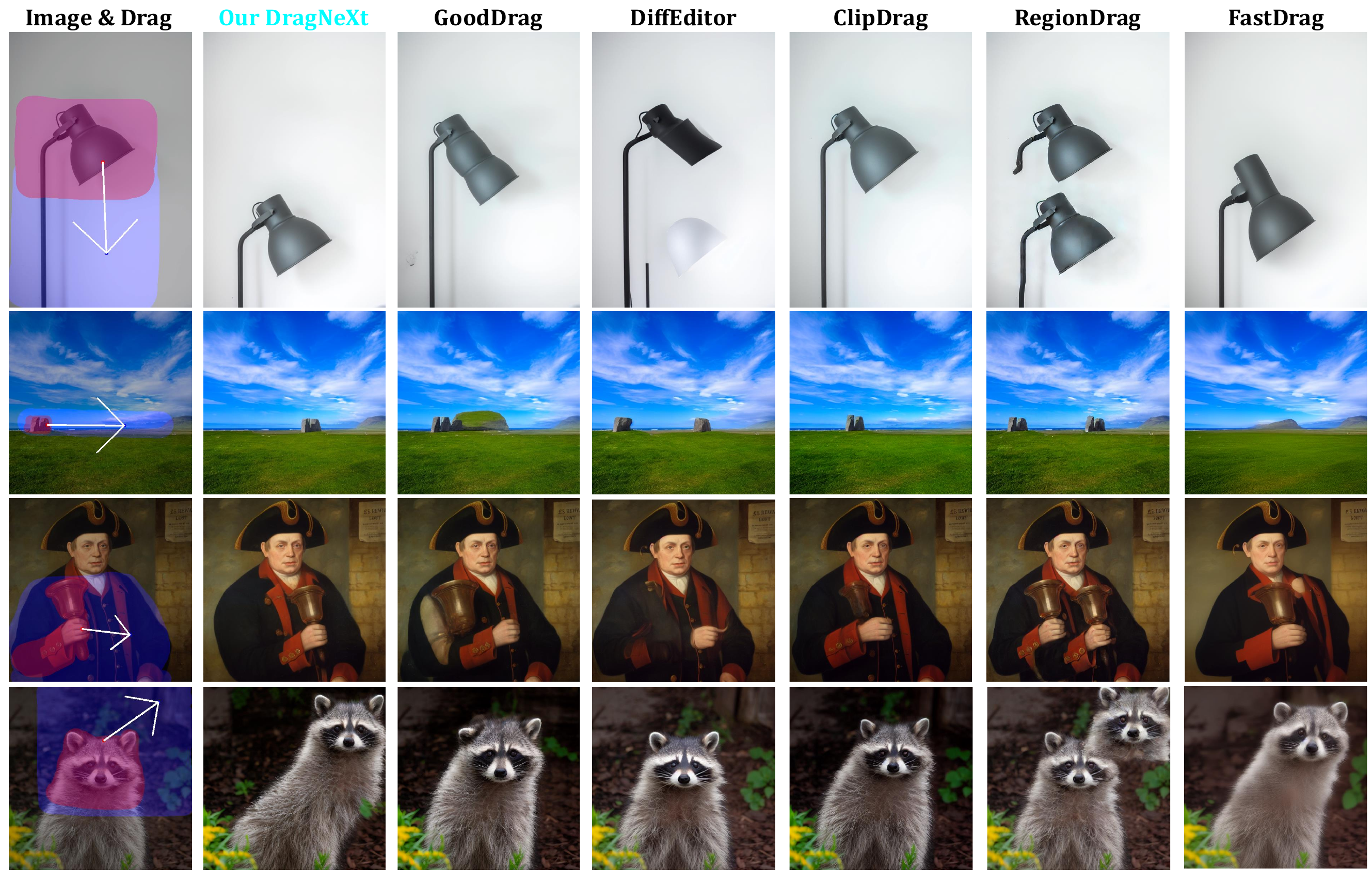}
\caption{\textbf{Experimental results on Drag-Based Image Editing at Relatively Long
Distances.}} 
\label{figure:15}
\end{figure*}

\begin{table}[t]
\centering
\includegraphics[width=0.6\linewidth]{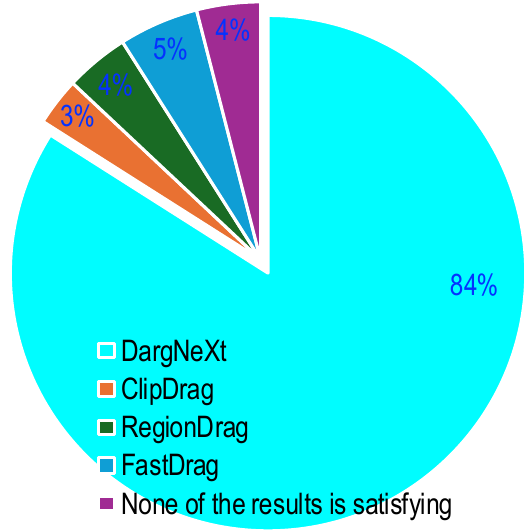}
\caption{\textbf{Voting results of our anonymous user study.}}
\label{figure:16}
\end{table}

\noindent
\textbf{REMARK 9. $\mapsto$Relationships between point-based 3D rotation and translation.} We argue there is no essential difference between point-based 3D rotation and translation, which both aim to drag handle points to the positions of target points. But translating the handle points of an object may lead to diffusion models to draw 3D rotation effects, e.g., the car’s 3D rotation can be realized by translating its front to the right as shown in the instruction in Figure \ref{figure:suppc}. \textbf{$\mapsto$Relationships between region-based 3D rotation and translation.} Similar to the point-based rotation used in GoodDrag, Region-based 3D rotation can be seen as translating an object's subregion.  

\section{E. More Experimental Results}

\subsection{Drag-Based Image Editing at Relatively Long Distances}
Dragging objects at a relatively long distance remains a major challenge in the current field of DBIE. Most of the existing methods only support dragging objects or regions over a short distance and are incapable of handling long-distance drag-based editing tasks. Although our method is not specifically designed for long-distance DBIE, we are surprised that it still exhibits superior performance compared to the recent counterparts. As exemplified in Figure \ref{figure:15}, we successfully drag the desk lamp, the stone, and the person's hand over a relative long distance while maintaining high editing quality. In contrast, the compared methods either fail to achieve long-distance dragging or to yield satisfactory quality. For instance, FastDrag easily causes unnatural deformation of objects, while RegionDrag is prone to resulting in artifacts in edited regions, as we mentioned in the main body of the paper. Also, ClipDrag, DiffEditor, and GoodDrag suffer from severe loss of regional details during long-distance dragging.

\noindent
\textbf{Why is DragNeXt superior?} We believe that the superiority of our method in long-distance dragging tasks lies in two aspects: (i) \textbf{DragNeXt} fully exploits region-level contextual information, effectively alleviating the influence of losing sparse handle points during long-distance dragging; and (ii) it incorporates progressive guidance from intermediate drag states, enabling a smoother and more stable dragging process. The strength in handling long-distance dragging tasks reveals that our method has great potential for achieving DBIE in complex scenarios. We plan to explore this point in our subsequent work.

\subsection{More Visualized Results}
\begin{figure*}[t!]
\centering
\includegraphics[width=0.8\linewidth]{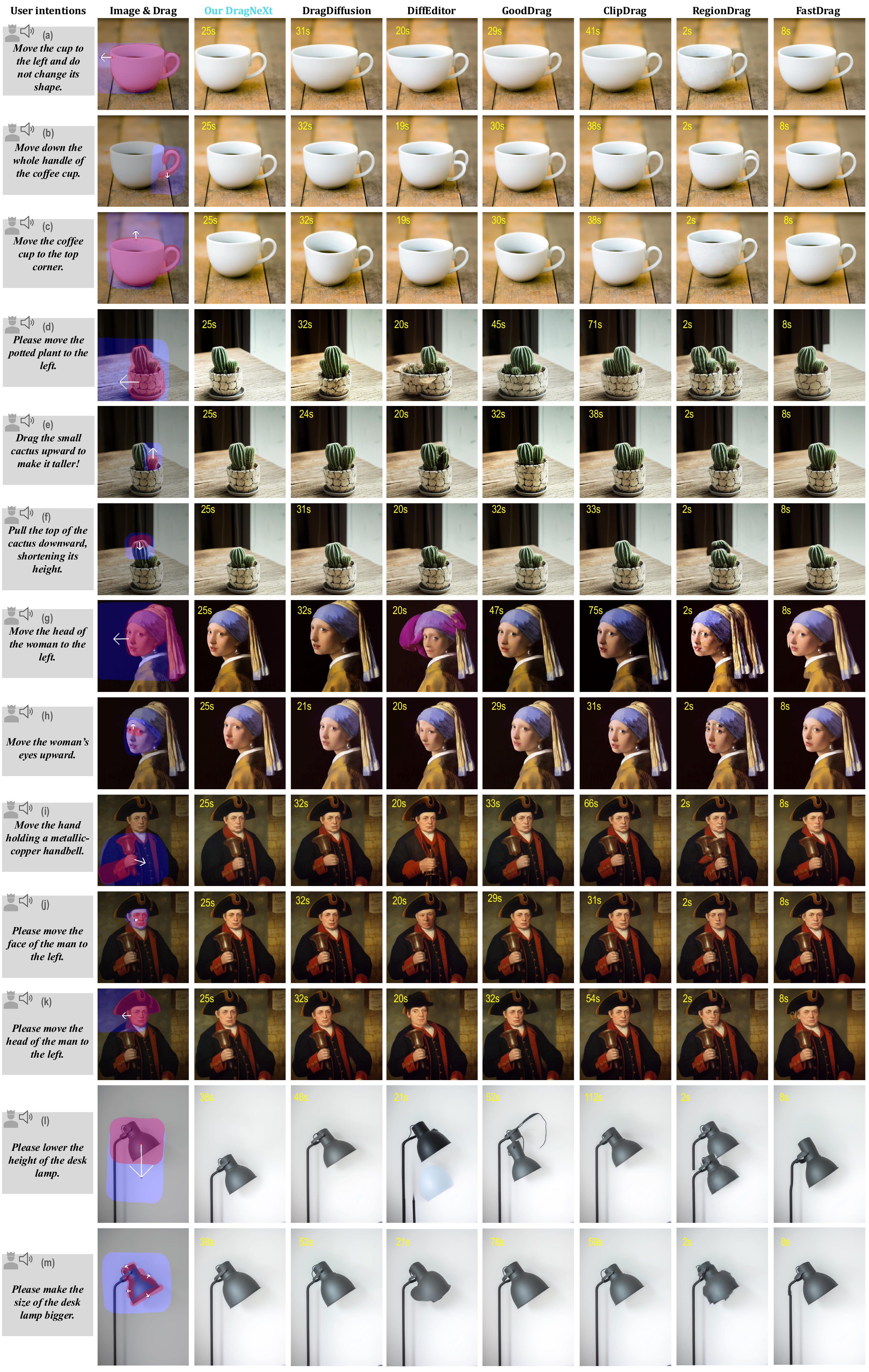}
\caption{\textbf{More experimental results---part I.}} 
\label{figure:12}
\end{figure*}

\begin{figure*}[t!]
\centering
\includegraphics[width=0.8\linewidth]{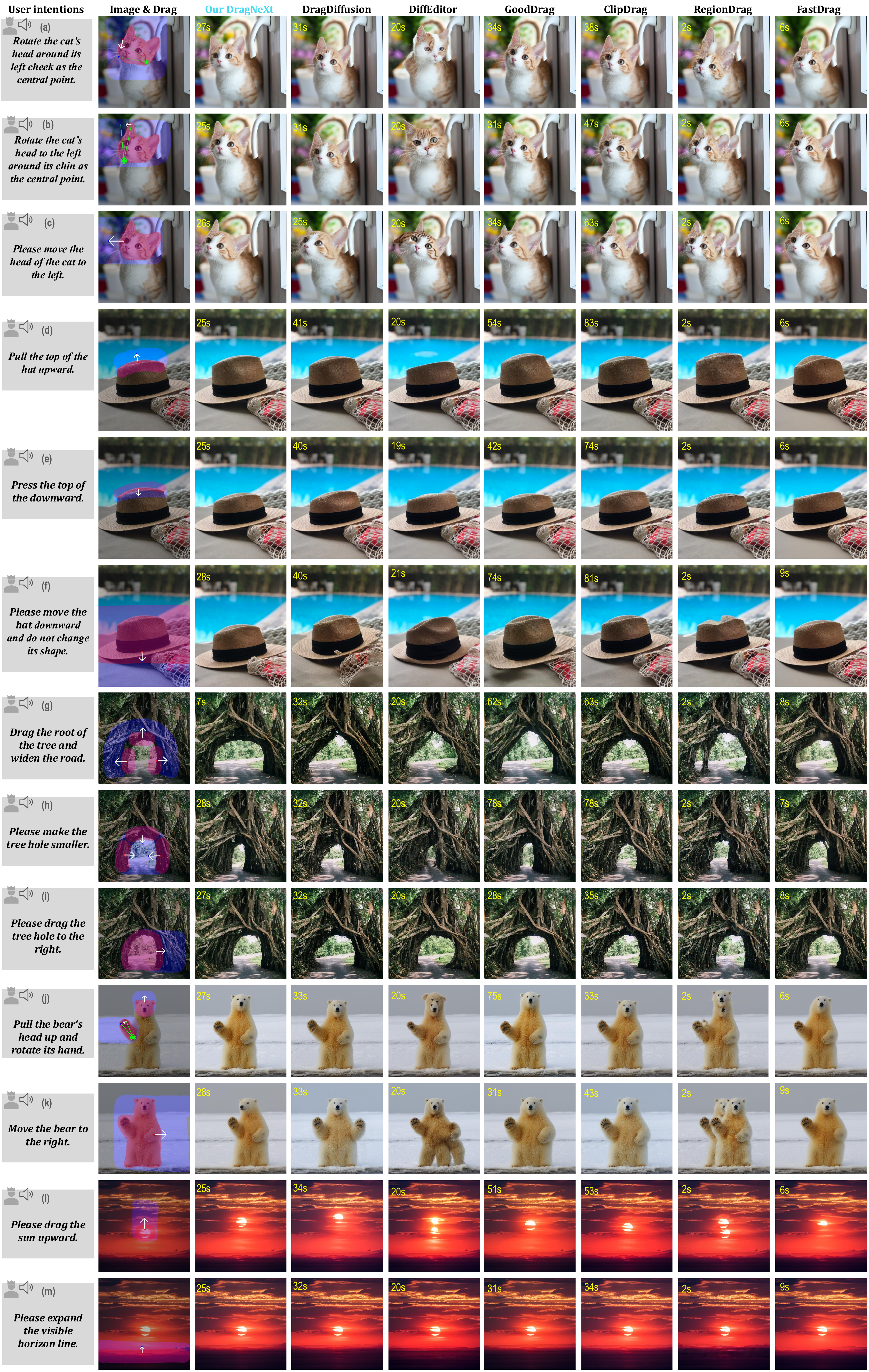}
\caption{\textbf{More experimental results---part II.}} 
\label{figure:13}
\end{figure*}

\begin{figure*}[t!]
\centering
\includegraphics[width=0.8\linewidth]{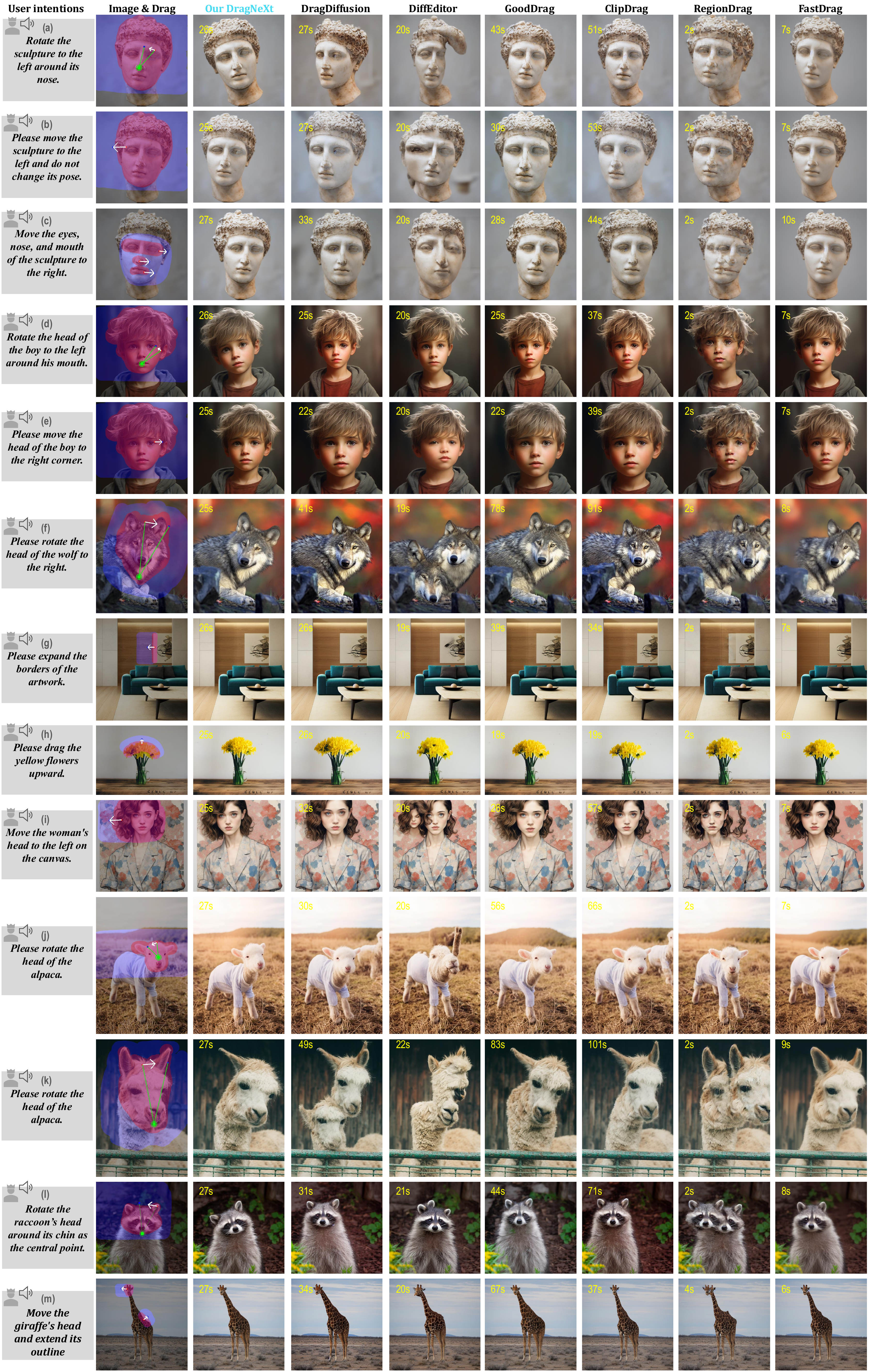}
\caption{\textbf{More experimental results---part III.}} 
\label{figure:14}
\end{figure*}

We provide more visualized results obtained by \textbf{DragNeXt} in Figures~\ref{figure:12}, \ref{figure:13}, and \ref{figure:14} as a supplementary for the experiments provided in the main body of the paper. These results lead to conclusions consistent with those discussed in Section~\ref{experiment}, further confirming the effectiveness of our approach in aligning with user intentions. For instance, in Figure~\ref{figure:12}(d), our method successfully moves the potted plant to the left, whereas all compared methods fail: DragDiffusion does not alter the plant’s position, and DiffEditor, GoodDrag, ClipDrag, RegionDrag, and FastDrag produce noticeable unnatural deformations. Moreover, our method also demonstrates superior performance in 2D rotation tasks, yielding more natural and visually consistent results than the compared counterparts, nearly all of which fail to generate satisfactory outcomes. We have also summarized our results in videos to provide a clearer and more intuitive illustration of the dragging effects yielded by our \textbf{DragNeXt}; for details, please kindly refer to the uploaded files in the supplementary material.


\begin{algorithm}[t]
    \small
    \caption{\textbf{Pseudocode of our proposed method}.}
    \label{algor:1}
    \SetAlgoLined
    \KwIn{an input image $\bm{x}$, user-specified handle regions $\bm{\mathcal{E}}=\{\bm{\vartheta}_i\}_{i = 1,\ldots,n}$ and drag instructions $\bm{\mathcal{C}}=\{\mathcal{T}_i,\bm{\mathcal{O}}_i\}_{i = 1,\ldots,n}$, hyperparamters $T$, $T'$, and $K$;}
    $\bm{z}_0 = \texttt{VAE\_Encoder}(\bm{x})$, $\{\bm{z}_1,\ldots,\bm{z}_T\} = \texttt{Inversion}(\bm{z}_0)$ \tcp*[r]{\textcolor{gray}{Encoding and inversion.}}
    \tcp{\textcolor{gray}{The denosing phase begins.}}
    \For{$t = T$ \KwTo $0$}{
        \If{$T' < t < T$}{
            $\bm{z}_t^0\leftarrow \bm{z}_t$; \\
            \tcp{\textcolor{gray}{Performing the PBSI strategy.}}
            \For{$k = 0$ \KwTo $K-1$}{
                $\{\bm{\rho}_{i}^{t,k},\bm{\Pi}_{\bm{\vartheta}_i\rightarrow\bm{\rho}_{i}^{t,k}}\}_{i = 1,\ldots,n}=\bigcup_{i = 1,\ldots,n} \delta (\bm{\vartheta}_i,\mathcal{T}_i,\bm{\mathcal{O}}_i,t,k)$;\
                $\mathcal{L}_{LRO}=\sum_{i = 1,\ldots,n}\|\mathcal{F}(\bm{z}_t^k)*\bm{\rho}_{i}^{t,k}-\mathcal{F}(\bm{z}_t).\texttt{copy.detach()}[\bm{\Pi}_{\bm{\vartheta}_i\rightarrow\bm{\rho}_{i}^{t,k}}]*\bm{\rho}_{i}^{t,k}\|_1+\mathcal{R}_{\bm{M}}$;\
                $\bm{z}_t^{k + 1} \longleftarrow\bm{z}_t^k - \frac{\partial\mathcal{L}_{LRO}}{\partial\bm{z}_t^k}$;\
            }
            $\bm{z}_{t - 1}=\bm{z}_{t}^{K-1}-\bm{\varepsilon}_{\bm{\Theta}}(\bm{z}_t^{K-1}; t, \bm{c})$;
        }
        \Else{$\bm{z}_{t - 1}=\bm{z}_{t}-\bm{\varepsilon}_{\bm{\Theta}}(\bm{z}_t; t, \bm{c})$; \tcp{\textcolor{gray}{Vanilla denoising.}}
        }
    }
    $\bar{\bm{x}}=\texttt{VAE\_Decoder}(\bm{z}_0)$; \tcp{\textcolor{gray}{Decoding latent embeddings.}} 
    \KwOut{an edited image $\bar{\bm{x}}$;}
\end{algorithm}


\section{F. Anonymous User Study}
\label{supp:user_study}
\begin{figure*}[t!]
\centering
\includegraphics[width=0.85\linewidth]{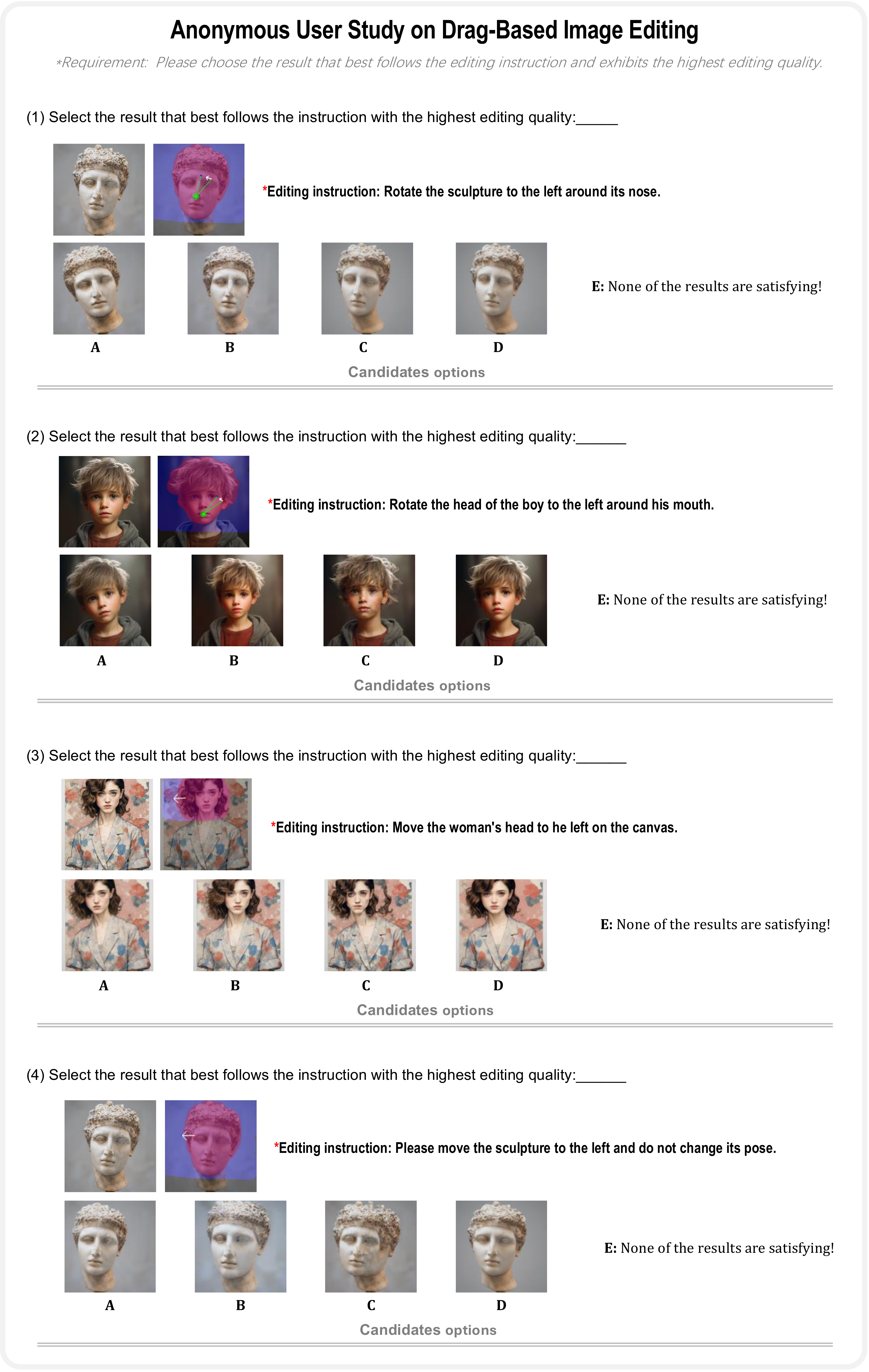}
\caption{\textbf{Questionnaire---Part I} \textit{(questions (1)$\sim$(4))}.} 
\label{figure:17}
\end{figure*}

\begin{figure*}[t!]
\centering
\includegraphics[width=0.85\linewidth]{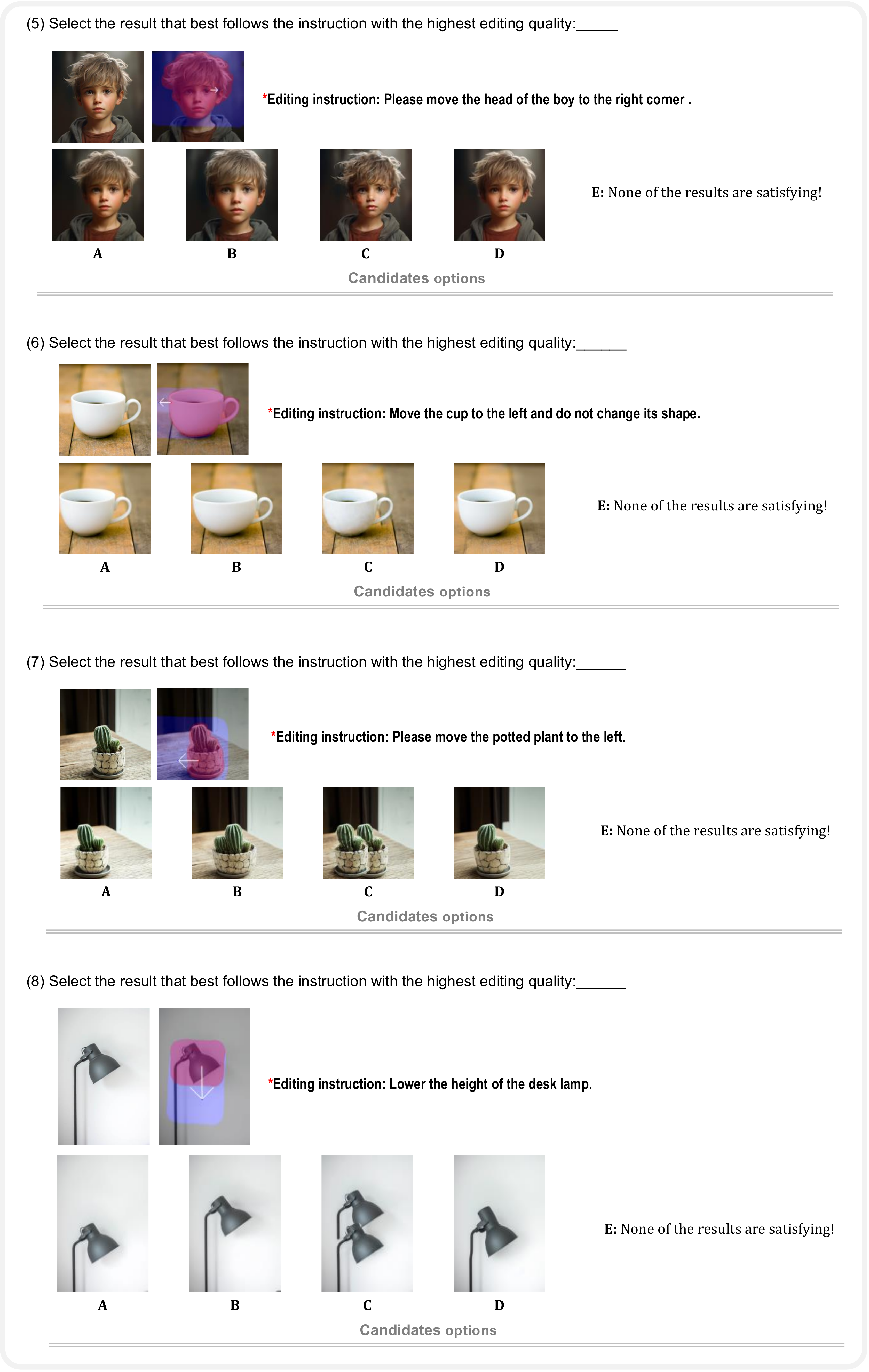}
\caption{\textbf{Questionnaire---Part II} \textit{(questions (5)$\sim$(8))}.} 
\label{figure:18}
\end{figure*}

\begin{figure*}[t!]
\centering
\includegraphics[width=0.85\linewidth]{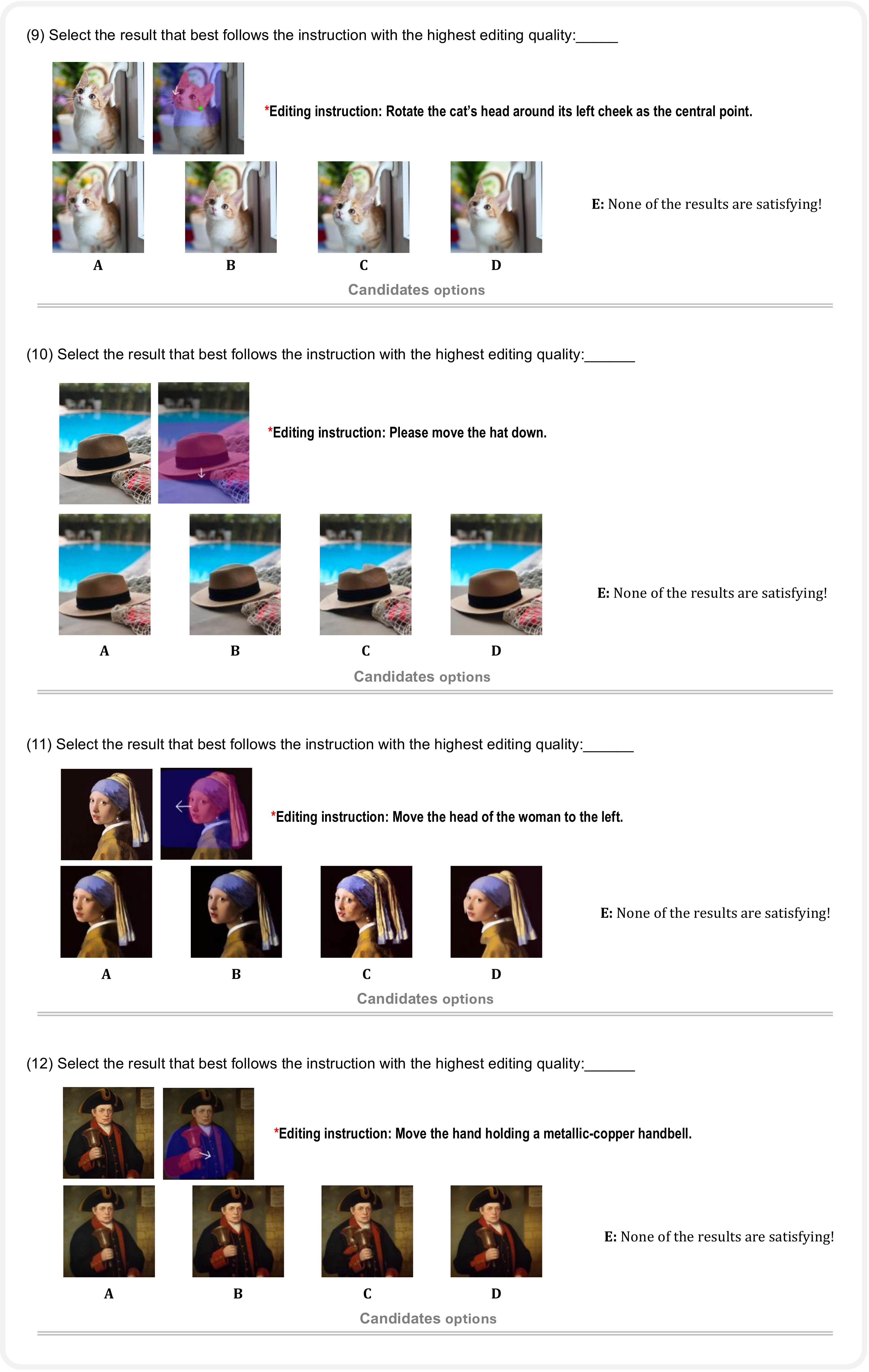}
\caption{\textbf{Questionnaire---Part III} \textit{(questions (9)$\sim$(12))}.} 
\label{figure:19}
\end{figure*}

\begin{figure*}[t!]
\centering
\includegraphics[width=0.85\linewidth]{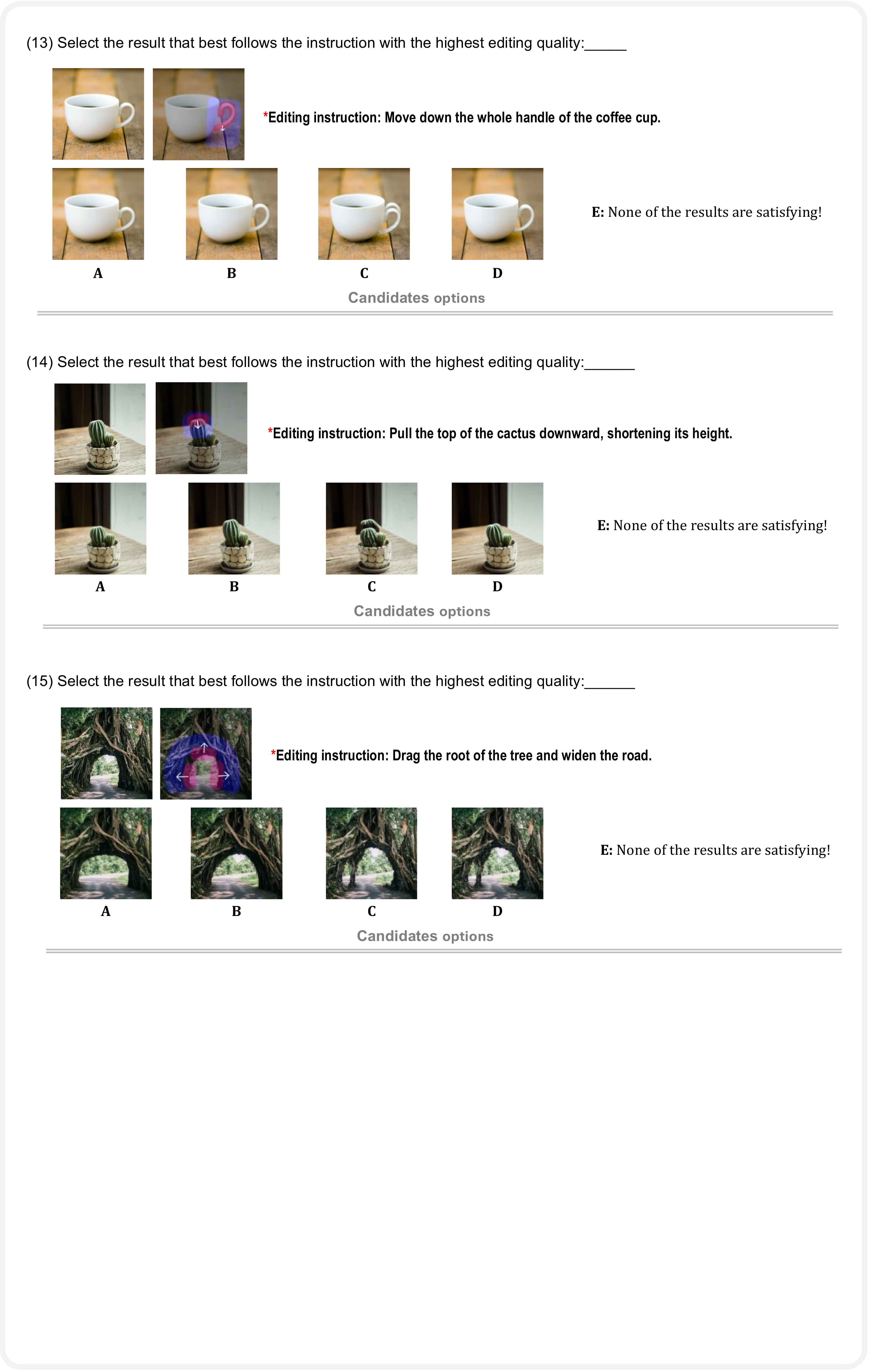}
\caption{\textbf{Questionnaire---Part IV} \textit{(questions (13)$\sim$(15))}.} 
\label{figure:20}
\end{figure*}

Since quantitative evaluation metrics may not fully demonstrate the effectiveness of our method in addressing the ambiguity issue and expectation-result misalignment, we additionally provide an anonymous user study, where a total of 26 participants are invited. The details about the questionnaire is summarized in Figure \ref{figure:17}, \ref{figure:18}, \ref{figure:19}, and \ref{figure:20}. The questionnaire totally consists of $15$ questions, where the $12$ items are closely related to the ambiguity issues mentioned in \textbf{Proposition~\ref{pro:ambiguity}}, and the $3$ items are used to assess the quality of edited images. Also, for each question, five candidate options are provided: 
\begin{itemize}
    \item the options \textbf{A---D} correspond to randomly ordered results generated by \textbf{\textcolor{SkyBlue}{DragNeXt}}, ClipDrag, RegionDrag, and FastDrag;
    \item the option \textbf{E} indicates that none of the results are satisfactory.
\end{itemize}
The reason for limiting the options to \textbf{A–E} is to reduce the participants’ workload, as selecting the most suitable result from too many options would be time-consuming and not user-friendly.
In Figure \ref{figure:16}, we provide the anonymous voting results from the invited participants. As can be seen from the figure, the voting results demonstrate the effectiveness of method again, e.g., the average results from the participants indicate that $84\%$ of our edited images are better than those of the compared models.

\noindent
\textbf{Why are ClipDrag, RegionDrag, and FastDrag Chosen as Compared Models?} These three methods are the most relevant to our research. ClipDrag \cite{CLIPDrag} addresses ambiguity in DBIE by incorporating textual guidance; in contrast, we reformulate DBIE as a Latent Region Optimization (LRO) problem to alleviate ambiguity while further improving the efficiency of the alternating workflow. \textbf{DragNeXt} advances RegionDrag and FastDrag by transforming forward, optimization-free manipulation into backward, self-interventional latent optimization, thereby fully leveraging the prior of pretrained diffusion models to avoid unrealistic and unnatural deterioration.

\section{G. Limitations and Future Work}
\label{supp:conclusion}
We summarize the pseudocode of our method in Algorithm~1. Here, we elaborate on the limitations of our current work, and introduce the corresponding plan for our future research.

\begin{itemize}
    \item \textbf{Limited types of used geometric transformation functions.} In our current work, we adopt two geometric transformations---translation and 2D rotation. Although these two transforms can realize translation, deformation, and 2D/3D rotation effects in DBIE, there may still exist some other useful transformation functions that have not yet been considered, such as scaling and shearing. In the future, we will explore more types of geometric transform functions. The main challenge of incorporating more geometric transformations lies in how to unify them into the current format of drag instructions. For example, dragging may lead not only to regular scaling but also to irregular or non-uniform scaling effects; however, properly defining scaling factors along each direction remains a non-trivial problem. 
    \item\textbf{Physically-driven drag effects.} Achieving physics-driven editing results remains a key challenging problem in DBIE. In this work, we extend point-guided dragging to a region-oriented geometric transformation paradigm to alleviate ambiguity and better leverage contextual information. In future work, we aim to realize physics-driven editing by incorporating physically based transformation functions.
    
    \item\textbf{Long-distance DBIE.} Our \textbf{DragNeXt} is not specifically designed for long-distance DBIE, despite achieving superior results in this challenging setting. In future work, we plan to conduct more experiments to identify the key factors for achieving high-quality results in long-distance DBIE and to further enhance our \textbf{DragNeXt}.

    \item\textbf{DBIE with Linear Attention}.  Diffusion models usually adopt standard self-attention with quadratic complexity, resulting in high computational costs. Recent advances in linear attention \cite{zhu2025dig,zhou2025care,liao2025vig} offer an effective way to alleviate this computational bottleneck. In future work, we plan to investigate incorporating linear attention into DBIE to further accelerate the editing process.

\end{itemize}


\end{document}